\documentclass[10pt,twocolumn,letterpaper]{article}

\usepackage{3dv}
\usepackage{times}
\usepackage{epsfig}
\usepackage{graphicx}
\usepackage{amsmath}
\usepackage{amssymb}
\usepackage{xcolor}
\usepackage{pict2e}
\usepackage[final]{changes}

\usepackage[pagebackref=true,breaklinks=true,letterpaper=true,colorlinks,bookmarks=false]{hyperref}

\renewcommand{\vec}[1]{\boldsymbol{#1}}
\newcommand{\mat}[1]{\mathbf{#1}}
\newcommand{\set}[1]{\mathcal{#1}}

\newcommand{\smpl}[0]{M}
\newcommand{\posefun}[0]{T}
\newcommand{\blendfun}[0]{W}
\newcommand{\offsetfun}[0]{B}
\newcommand{\jointfun}[0]{J}

\newcommand{\pose}[0]{\vec{\theta}}
\newcommand{\shape}[0]{\vec{\beta}}

\newcommand{\blendweights}[0]{\mat{W}}
\newcommand{\template}[0]{\mat{T}}

\newcommand{\vertex}[0]{\vec{v}}
\newcommand{\normal}[0]{\vec{n}}
\newcommand{\normals}[0]{\mat{N}}
\newcommand{\shadingoffset}[0]{s}
\newcommand{\shadingoffsets}[0]{\mat{s}}
\newcommand{\ray}[0]{\mat{r}}
\newcommand{\raycomp}[0]{\vec{r}}

\newcommand{\offsets}{\mat{D}}

\newcommand{\smplfine}[0]{\smpl_f}
\newcommand{\edge}{\set{E}}

\newcommand{\topology}{\set{N}}
\newcommand{\neighbors}{\set{N}}
\newcommand{\landmarkmatches}{\set{L}}
\newcommand{\landmark}{\vec{l}}

\newcommand{\sh}[0]{\vec{c}}
\newcommand{\projection}[0]{\mat{P}}
\newcommand{\image}[0]{I}
\newcommand{\visvertices}[0]{\set{V}}

\threedvfinalcopy % *** Uncomment this line for the final submission

\def\threedvPaperID{13} % *** Enter the 3DV Paper ID here

\ifthreedvfinal\fi

\begin{document}
	
	%%%%%%%%% TITLE
	\title{Detailed Human Avatars from Monocular Video}
	
	\author{Thiemo Alldieck\textsuperscript{1,2}\hspace{-3.2mm}
		%{\tt\small alldieck@cg.cs.tu-bs.de}
		\and
		Marcus Magnor\textsuperscript{1}\hspace{-3.2mm}
		%{\tt\small magnor@cg.cs.tu-bs.de}
		\and
		Weipeng Xu\textsuperscript{2}\hspace{-3.2mm}
		%{\tt\small xu@mpi-inf.mpg.de}
		\and
		Christian Theobalt\textsuperscript{2}\hspace{-3.2mm}
		%{\tt\small theobalt@mpi-inf.mpg.de}
		\and
		Gerard Pons-Moll\textsuperscript{2}
		%{\tt\small gpons@mpi-inf.mpg.de}
	}
	
	\makeatletter
	\let\@oldmaketitle\@maketitle% Store \@maketitle
	\renewcommand{\@maketitle}{
		\@oldmaketitle% Update \@maketitle to insert...
		\centering
		\vspace{-10mm}
		{\small \textsuperscript{1}Computer Graphics Lab, TU Braunschweig, Germany}\\
		{\small	\textsuperscript{2}Max Planck Institute for Informatics, Saarland Informatics Campus, Germany}\\
		{\tt\scriptsize \{alldieck,magnor\}@cg.cs.tu-bs.de \{wxu,theobalt,gpons\}@mpi-inf.mpg.de}\\
		\vspace{4mm}
		\includegraphics[width=0.9\textwidth]{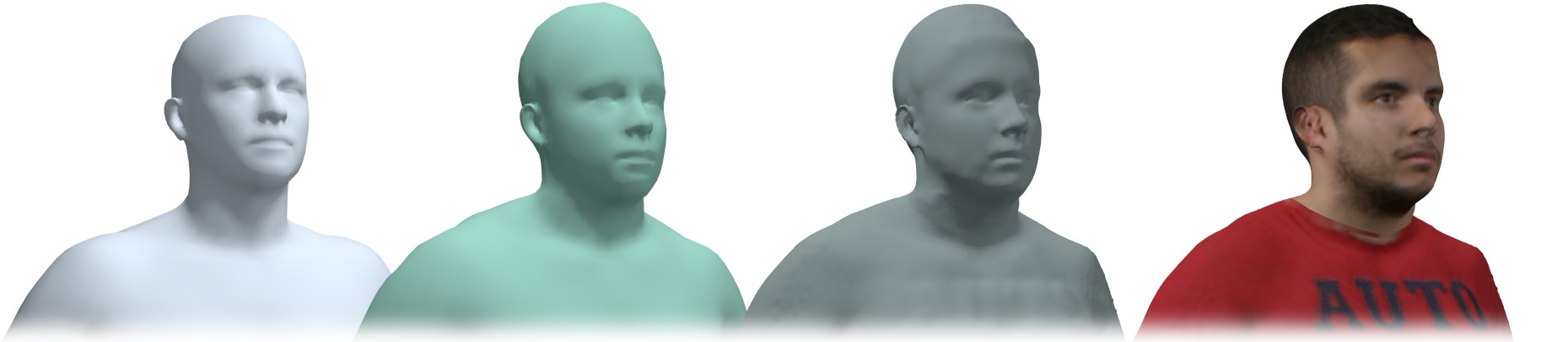}\\
		\refstepcounter{figure}\normalfont Figure~\thefigure: Our method 
		%calculates 
		creates a detailed avatar
		%subject 
		from a monocular video of a person
		%rotating 
		turning around. Based on the SMPL model, we first compute a medium-level avatar, then add subject-specific details and 
		%determine 
		finally generate
		a seamless texture. 
		\label{fig:teaser}
		\vspace{4mm}
	}
	\makeatother
	
	\maketitle
	% \thispagestyle{empty}
	
	%%%%%%%%% ABSTRACT
	\begin{abstract}
		%\vspace{-2mm}
		We present \deleted{the first} \added{a novel} method for \added{high} detail-preserving human avatar creation from monocular video. A parameterized body model is 
		%extended 
		refined   
		and optimized to maximally resemble subjects from a video showing them from all sides. 
		%The calculated 
		Our avatars feature a natural face, 
		%hair, 
		hairstyle, clothes with garment wrinkles, and high-resolution texture. 
		%This work introduces 
		Our paper contributes facial landmark and shading-based human body shape refinement, %the creation of
		a semantic texture prior, and a novel texture stitching strategy, resulting in the most sophisticated-looking 
		%method to obtain 
		human avatars obtained from a single video to date. Numerous results show the 
		%effectiveness 
		robustness and versatility of our method\deleted{, and a}\added{. A} user study illustrates its superiority
		%clear improvement 
		over the state-of-the-art in terms of identity preservation, level of detail, realism, and overall user preference.
	\end{abstract}
	\vspace{-5mm}
	
	%%%%%%%%% BODY TEXT
	\section{Introduction}
	\label{sec:introduction}
	The automatic generation of personalized 3D human models is needed for many applications, including virtual and augmented reality, entertainment, teleconferencing, virtual try-on, biometrics or surveillance. A personal 3D human model should comprise all the details that make us different from each other, such as hair, clothing, facial details and shape. Failure to faithfully recover all details results in users not feeling identified with their self-avatar. 
	
	To address this challenging problem, researchers have used very expensive recording equipment including 3D and 4D scanners~\cite{pons2015dyna,dfaust:CVPR:2017,smpl2015loper} or multi-camera studios with controlled lighting~\cite{robertini2017multi, leroy2017multi}. 
	An alternative is to use passive stereo reconstruction~\cite{fuhrmann2014mve,Newcombe2011DTAM} with a camera moving around the person, but the person has to maintain a static pose which is not feasible in practice.
	Using depth data as input, the field has seen significant progress in reconstructing accurate 3D body models~\cite{Bogo:ICCV:2015,weiss2011home,zhang2014quality} or free-form geometry~\cite{zollhofer2014real,newcombe2015dynamicfusion,orts2016holoportation,dou2016fusion4d} or both jointly~\cite{DoubleFusion2018}. Depth cameras are however much less ubiquitous than RGB cameras. 
	
	Monocular RGB methods are typically restricted to prediciting the parameters of a statistical body model~\cite{omran2018neural,kanazawa2018endtoend,pavlakos2018humanshape,bogo2016smplify,bualan2008naked,hasler2010multilinear}. 
	To the best of our knowledge, the only exception is a recent method~\cite{alldieck2018video} that can reconstruct shape, clothing and hair geometry from a monocular video sequence of a person rotating in front of the camera. The basic idea
	is to fuse the information from frame-wise silhouettes into a canonical pose, and optimize a free-form shape regularized by the SMPL body model~\cite{smpl2015loper}. While this is a significant step in 3D human reconstruction from monocular video, the reconstructions are overly smooth, lack facial details and the textures are blurry. This results in avatars that do not fully retain the identity of the real subjects. 
	
	In this work, we extend~\cite{alldieck2018video} in several important ways to improve the quality of the 3D reconstructions and textures. Specifically, we incorporate information from facial landmark detectors, shape-from-shading, and we introduce a new algorithm to efficiently stitch partial textures coming from frames of the moving person. Since the person is moving, information (projection rays from face landmarks and normal fields from shading cues) can not be directly fused into a single reconstruction. Hence, we track the person's pose using SMPL~\cite{smpl2015loper}; then we apply an inverse pose transformation to frame-wise projection rays and normal fields to fuse all the evidence in a canonical T-pose; in that space, we optimize a high-resolution shape regularized by SMPL. 
	Precisely, with respect to previous work, our approach differs in four important aspects that allow us better preserve subject identity and details in the reconstructions:
	
	\textbf{Facial landmarks:} Since the face is a crucial part of the body, we incorporate \added{2D facial landmark detections} \deleted{facial landmarks} into the 3D reconstruction objective. \deleted{Facial landmarks are detected frame-wise using a state-of-the-art CNN based detector~\cite{simon2017hand}. The projection ray going from the center of the camera through the detection should intersect the corresponding landmark on the posed model.
		To fuse the temporal detections, we unpose the projection rays at every frame; then in a canonical pose, we minimize the corresponding model point to ray distances.}\added{To gain robustness against misdetections, we fuse temporal detections by transforming the landmark projection rays into the joint T-pose space.}  \deleted{Since we integrate rays from every frame, reconstructions are robust to \deleted{miss-detections} \added{misdetections} and noise.}  
	
	\textbf{Illumination and shape-from-shading:} Shading is a strong cue to recover fine details such as wrinkles. Most shape-from-shading approaches focus on adding detail to static objects. Here, we perform shape-from-shading at every frame, obtaining frame-wise partial 3D normal fields \added{that are then fused in T-pose space for final reconstruction.} \deleted{; for every incoming frame, we unpose the 3D normal fields, and optimize the vertices of the 3D reconstruction to match the unposed normals.}  
	
	\textbf{Efficient texture stitching:} \deleted{Seamless stitching of partial textures from different camera views is challenging even for static objects. For moving articulated objects the problem is even more difficult.} \added{Seamless stitching of partial textures from different camera views is particulary hard for moving articulated objects.}\deleted{Typical problems are artifacts at the boundary regions due to \deleted{miss-registrations} \added{misregistrations}. One option is to compute the mean or the median
		% deleted	
		%~\cite{bernardini2001high, debevec1996modeling, Ofek:1997:MTI:616045.618417, pighin2006synthesizing}
		of partial textures from every frame, which results in blurry textures. A more sophisticated alternative is to assign the RGB value of one the views to each texture pixel (texel), while preserving spatial smoothness.}
	\added{To prevent blurry textures, one typically assigns the RGB value of one the views to each texture pixel (texel), while preserving spatial smoothness.}
	Such assignment problem can be formulated as a multi-labeling assignment, where number possible labels grows with the number of views. Consequently, the computational time and memory becomes intractable for \deleted{our problem where we merge $60$ frames. Hence, we rephrase the problem into a binary labeling problem} \added{for a large number of labels} -- we define a \added{novel} \emph{texture update energy function} which can be minimized \added{efficiently} with a graph cut for every new incoming view. \deleted{Specifically, we decide for every texel whether it should be updated with the incoming view or not. This results in sharp textures with less artifacts at the boundaries.}
	
	\textbf{Semantic texture stitching:} Aside from stitching artifacts, texture spilling is another common problem. For example texture that corresponds to the clothing often floods into the skin region. 
	To minimize spilling we add an additional semantic term into the texture update energy. The term penalizes updating a texel with an RGB value that is unlikely under a part-based appearance distribution.
	\deleted{To obtain part appearance models, we compute a semantic segmentation~\cite{gong2017look} of the person for every frame and fuse it into the texture (Fig.~\ref{fig:semantictexture}).
		This allows us to compute per part appearance models using a Gaussian mixture model.} This \deleted{additional} semantic \deleted{segmentation} appearance term significantly reduces \deleted{texture} spilling, and implicitly "connects" texels belonging to the same part.

	The result is the most sophisticated method to obtain detailed 3D human shape reconstructions from single monocular video. Since metric based evaluations such as scan to mesh distances do not reflect the perceptual quality, we performed a user study to assess the improvement of our method. The results show that users prefer our avatars over state-of-the-art $89.64 \%$ of the times and they think our reconstructions are more detailed $95.72 \%$ of the times.

	%\subsection*{Novelity of the paper}
	%
	
	%
	%Both claims should be supported by the user study.
	
	%\subsubsection*{Technical}
	
	%\begin{itemize}
	%	\item Integration of facial features into the CVPR pipeline
	%	\item Integration of shape-from-shading into the CVPR pipeline via normal unposing
	%	\item Integration of matches between frames into the CVPR pipeline for better regularization (?)
	%	\item Stitching a segmentation texture as prior step for texture generation; fit GMM per segment as color prior
	%	\item Stitching the texture on texel level (not faces)
	%	\item New texture stitching algorithm transforming multi-label graph cut into binary decisions
	%\end{itemize}
	
	%\subsubsection*{Minor?}
	
	%\begin{itemize}
	%	\item Extended SMPL formulation
	%	\item Using a CNN predict shading for shape-from-shading
	%	\item Warp and flow based matching to make optical flow capture details (?)
	%	\item Structural dissimilarity for preserving facial expression while texture stitching
	%\end{itemize} 
	
	\section{Related work}
	\label{sec:related}

	\textbf{Modeling the human body} is a long-standing problem in computer vision.
	%Previous work on this topic can be grouped into two categories: (1) modeling a static model (2) capture the dynamically deforming body shape.
	%Our method falls into the first category.
	Given a densely distributed multi-camera system, one can make use of multi-view stereo methods~\cite{koch1998multi} for reconstructing the human body~\cite{fuhrmann2014mve,galliani2015massively,jancosek2011multi,zheng2014patchmatch}. More advanced systems allow reconstruction of body shape under clothing~\cite{zahng2017shapeundercloth,yang2016estimation,wuhrer2014estimation}, joint shape, pose and clothing reconstruction~\cite{ponsmoll2017clothcap}, or capture body pose and facial expressions~\cite{joo2018total}.
	However, such setups are expensive and require complicated calibration.
	
	Hence, monocular 3D reconstruction methods~\cite{newcombe2011kinectfusion,Newcombe2011DTAM} are appealing but require depth images from many view points around a static object and humans can not hold a static pose for a long time.
	Therefore, nonrigid deformation of the human body has to be taken into account.
	%Most of such existing methods are based on depth sensors.
	%So either the subject needs to stand on a turntable or the sensor rotates around the subject.
	%The observation from all view points are merged using method like~\cite{newcombe2011kinectfusion}.
	%Similar online fusion methods are also available for color cameras~\cite{Newcombe2011DTAM}.
	Many methods are based on depth sensors and require the subject to hold the same pose.
	For example, in~\cite{3Dportraits,cui2012kinectavatar,shapiro2014rapid,zeng2013templateless}, the subject alternatively makes a certain pose and rotates in front of the sensor.
	Then, several depth snapshots taken from different view points are fused to generate a complete 3D model. 
	Similarly,~\cite{tong2012scanning} proposes to use a turntable to rotate the subject to minimize pose variations. 
	In contrast, the methods of \cite{Bogo:ICCV:2015,weiss2011home,zhang2014quality} allow a user to move freely in front of the sensor.
	%The main idea is to fuse the depth measurement undergoing nonrigid deformation.
	In recent years, real time nonrigid depth fusion has been achieved~\cite{newcombe2015dynamicfusion,innmann2016volume,slavcheva2017killingfusion}.
	These methods usually maintain a growing template and consist of two alternating steps, i.e. a registration step, where the current template is aligned to the new frame, and a fusion step, where the observation in the new frame is merged to the template.
	%Comparing to template based methods, they are less robust.
	However, these methods typically suffer from ``phantom surfaces'' artifacts during fast motion.
	In~\cite{DoubleFusion2018}, this problem is alleviated by using SMPL to constraint tracking.
	Model based monocular methods~\cite{bogo2016smplify,dibra2017human,guan2009estimating,bualan2008naked,hasler2010multilinear,Pons-Moll_MRFIJCV,ponsmollModelBased} have recently been integrated with deep learning~\cite{omran2018neural,kanazawa2018endtoend,pavlakos2018humanshape}.
	However, they are restricted to predicting the parameters of a statistical body model~\cite{smpl2015loper,anguelov2005scape,hasler2009statistical,zuffi2015stitched,pons2015dyna}. 
	There are two exceptions, that recover clothing and shape from a single image~\cite{guo2012clothed,chen2013deformable} but these methods require manual initialization of pose and clothing parameters.
	%As exceptions, shape and clothing is recovered from a single image in~\cite{guo2012clothed,chen2013deformable} but the user needs to click points in the image and select the clothing types from a database.
	\cite{alldieck2018video} is the first method capable of reconstructing full 3D shape and clothing geometry from a single RGB video. Users can freely rotate in front of the camera while roughly holding the A-pose.
	Unfortunately, this approach is restricted to recover only medium-level details.
	The fine-level details such as garment wrinkles, subtle geometry on the clothes and facial features, which are essential elements for preserving the identity information, are missing.
	Our goal is to recover the missing fine-level details of the geometry and improve the texture quality such that the appearance identity information can be faithfully recovered.
	
	Another branch of work in human body reconstruction is more focused on capturing the dynamic motion of the character.
	Works either recover articulated skeletal motion~\cite{stoll2011fast,mehta2017vnect,gavrila1996,sigal2004tracking,alldieck2017optical,huang2017towards}, or surfaces with deformed clothing, usually called performance capture.
	In performance capture many approaches reconstruct a 3D model for each individual frame~\cite{starck2007surface,inria_2017,collet2015high} or fuse a window of frames~\cite{orts2016holoportation,dou2016fusion4d}.
	However, these methods cannot generate a temporal coherent representation of the model, which is an important characteristic for many applications.
	To solve this, methods register a common model to results of all frames~\cite{cagniart_meshdeform}, use volumetric representation for surface tracking~\cite{InriaVolumetric_2015,huang2016volumetric}, or assume a pre-built static template.
	Again, most of those methods are based on multi-view images~\cite{deAguiar2008performance,gall2009motion,plankers2001articulated,carranza2003free,rhodin2016general,robertini2017multi}.
	There are attempts on reducing the number of cameras, such as the stereo method~\cite{wu2013set}, single view depth based method~\cite{zollhofer2014real} and the recent monocular RGB based method~\cite{MonoPerfCap_SIGGRAPH2018}.
	Note that the result of our method can be used as the initial template for above-mentioned template based performance capture methods.

	\textbf{Shape-from-shading} is also highly related to our method.
	A comprehensive survey can be found in~\cite{zhang1999shape}.
	We only discuss the application of shape-from-shading in the context of human body modeling.
	Geometric details, e.g. folds in the non-textured region, are difficult to capture with silhouette or photometric information.
	%When using a complex controlled light stage indoors, more surface
	%detail can be reconstructed by exploiting visible shading information~\cite{vlasic2009dynamic}.
	In contrast, shape-from-shading captures such details~\cite{wu2013set,wu2011shading,haefner2018fight}.
	%Used to refine 3D reconstruction from depth ~\cite{zollhofer2015shading}.
	There are also approaches for photometric stereo which recover the shape using controlled light stage setup~\cite{vlasic2009dynamic}.
	
	\textbf{Texture generation} is an essential task for modeling a realistic virtual character, since a texture image can describe the material properties that cannot be modeled by the surface geometry.
	The key of a texture generation method is how to combine texture fragments created from different views.
	Many early works blend the texture fragments using weighted averaging across the entire surface~\cite{bernardini2001high, debevec1996modeling, Ofek:1997:MTI:616045.618417, pighin2006synthesizing}.
	Others make use of mosaicing strategies, which yields sharper results~\cite{baumberg2002blending,lensch2001silhouette,niem1997automatic,rocchini1999multiple}.
	\cite{lempitsky2007seamless} is the first to formulate texture stitching as a graph cut problem.
	Such formulation has been commonly used in texture generation for multi-view 3D reconstruction.
	However, without accurately reconstructed 3D geometry and registered images, these methods usually suffer from blurring or ghosting artifacts.
	To this end, many methods focus on compensating registration errors~\cite{eisemann2008floating,bi2017patch,waechter2014let,FuYanYangEtAl,zhou2014color}.
	In our scenario, the registration misalignment problem is even more severe, due to our challenging monocular nonrigid setting.
	Therefore, we propose to take advantage of semantic information to better constrain our problem.
	\deleted{Furthermore, in order to obtain a sharp and seamless texture, we propose a novel approach to solve the texture stitching on a per texel level, in contrast to that on a per face level as in other works.}
	
	\section{Method}
	\label{sec:method}
\begin{figure}
		\centering
		\includegraphics[width=1\linewidth]{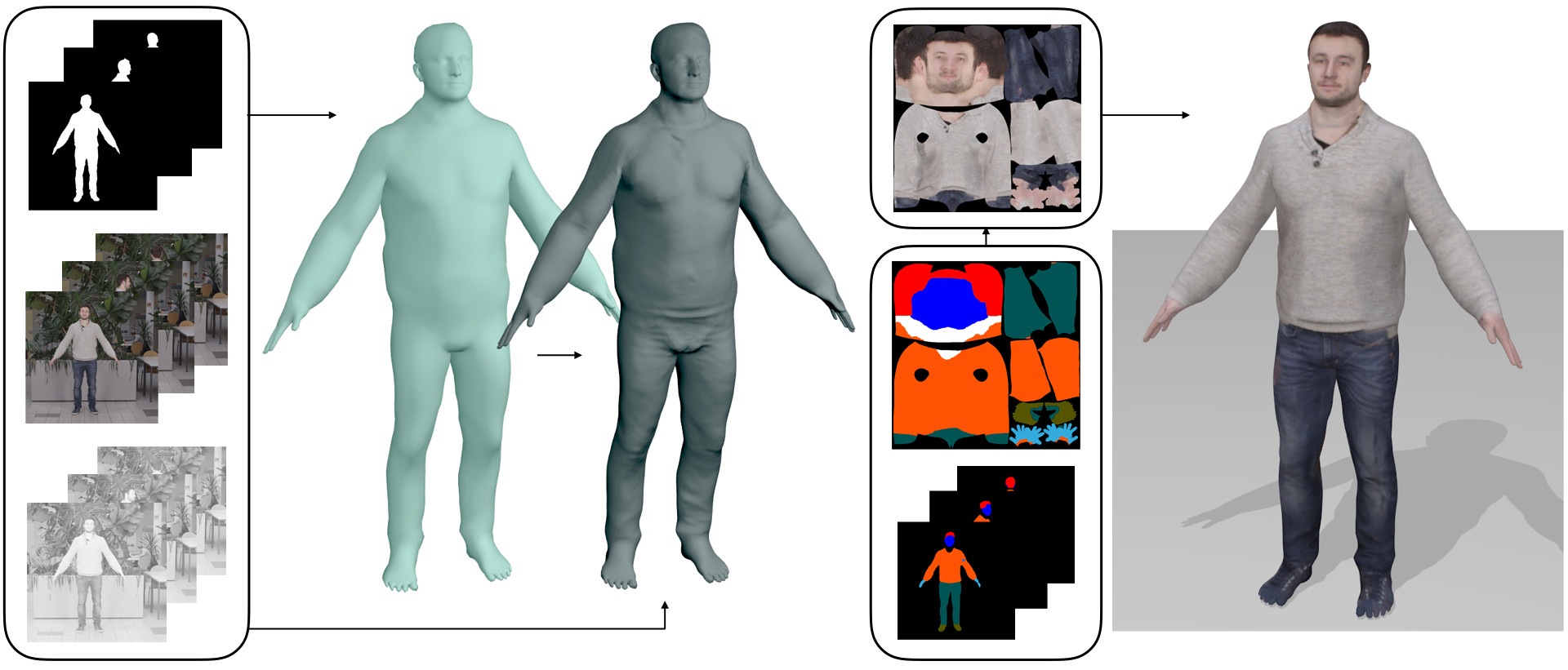}
		\begin{minipage}{1.0\textwidth}
			\footnotesize
			\vspace{-6mm}
			\hspace{23mm}a)\hspace{10mm}b)\hspace{32.5mm}c)
		\end{minipage}
		\vspace{-7mm}
		\caption{Our method 3-step method: We first estimate a medium level body shape based on segmentations (a), then we add details using shape-from-shading (b). Finally we compute a texture using a semantic prior and a novel graph cut optimization strategy (c).}
		\label{fig:pipeline}
		\vspace{-1mm}
	\end{figure}

	In this paper, our goal is to create a detailed avatar from an RGB video of a subject rotating in front of the camera.
	The focus lies hereby on fine-level details, that model a subject's identity and individual appearance.
	As shown in Fig.~\ref{fig:pipeline}, our method reconstructs a textured mesh model in a coarse-to-fine manner, which consists of three steps: First we estimate a rough body shape of the subject, similar to \cite{alldieck2018video}, where the medium-level geometry of the clothing and skin is reconstructed.
	Then we add fine-level geometric details, such as garment wrinkles and facial features, based on shape-from-shading.
	Finally, we compute a seamless texture to capture the texel-level appearance details.
	In the following, we first describe our body shape model, and then discuss the details of our three steps.
	
	\subsection{Subdivided SMPL body model}
	
	Our method is based on the SMPL body model~\cite{smpl2015loper}. 
	However, the original SMPL model is too coarse to model fine-level details such as garment wrinkles and fine facial features.
	To this end, we adapt the model as follows.
	
\begin{figure}
		\centering
		\begin{minipage}{0.55\linewidth}
			\setlength{\unitlength}{0.1\linewidth}
			\begin{picture}(10, 5.8)
			\put(0,0){\includegraphics[width=\linewidth]{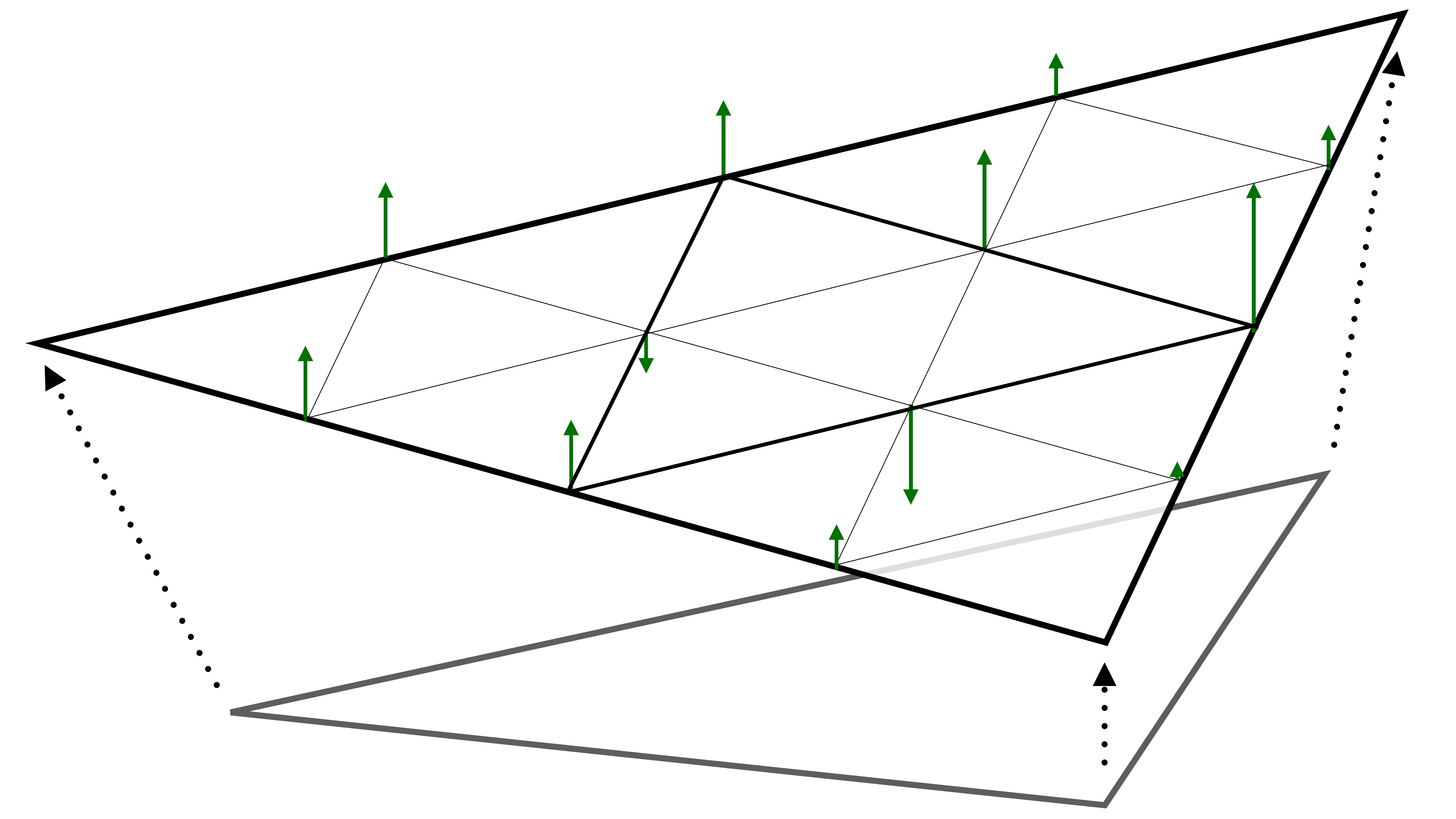}}
			\put(1.5,1.3){$\vec{d_i} \in \offsets$}
			\put(5.1,5){$s_e\normal_e$}
			\end{picture}
		\end{minipage}
		\vspace{1mm}
		\caption{One face of the new SMPL formulation. The displacement field vectors $\vec{d_\ast}$ and the normal displacements $s_\ast\normal_\ast$ form the subdivided surface.}
		\label{fig:sfssmpl}
		\vspace{-1mm}
	\end{figure}

	The SMPL model is a parameterized human body model described by a function of pose $\pose$ and shape $\shape$ returning $N=6890$ vertices and $F=13776$ faces.
	As SMPL only models naked humans, we use the extended formulation from~\cite{alldieck2018video} allowing offsets $\offsets$ from the template $\template$:
	\begin{equation}
	\smpl(\shape,\pose,\offsets) = \blendfun(\posefun(\shape,\pose,\offsets), \jointfun(\shape), \pose, \blendweights)
	\label{eq_smpl_coarse}
	\end{equation}
	\begin{equation}
	\posefun(\shape,\pose,\offsets) = \template + \offsetfun_s(\shape) + \offsetfun_p(\pose) + \offsets
	\end{equation}
	where $\blendfun$ is a linear blend-skinning function applied to a rest pose $\posefun(\shape,\pose,\offsets)$ based on the skeleton joints $\jointfun(\shape)$  and after pose $\offsetfun_p(\pose)$ and shape dependent $\offsetfun_s(\shape)$ deformations. The inverse function $\smpl^{-1}(\shape,\pose,\offsets)$ \emph{unposes} the model and brings the vertices back into the canonical T-pose.
	%This formulation allows to explain medium-level clothing and hair properties.
	As we aim for fine details and a subject's identity, we further extent the formulation. 
	As shown in Fig.~\ref{fig:sfssmpl}, we subdivide every edge of the the SMPL model twice. Every new vertex is defined as:% and model every new vertex as a function of the two neighboring vertices along the subdivided egde and an offset $\shadingoffset$ in normal direction $\normal$:
	\begin{equation}
	\small
	\vertex_{N+e} = 0.5(\vertex_i + \vertex_j) + \shadingoffset_{e}\normal_{e}, \quad (i, j) \in \edge_e
	\end{equation}
	%\begin{equation}
	%\small
	%\normal_{e} = \frac{\normal_j + \normal_k}{||\normal_j|| + ||\normal_k||}, \quad (i, j) \in \edge_e
	%\end{equation}
	%
	where $\edge$ defines the pairs of vertices forming an edge and $\normal_{e}$ is the average normal between the normals of the vertex pair.
	$\shadingoffset \in \shadingoffsets$ defines the displacement in normal direction $\normal_{e}$. $\normal_{e}$ is calculated at initialization time in unposed space and can be posed according to $\blendfun$. %\TA{Do we really new $\blendfun$ for posing? Can't we calculate the direction based on the supporting vertices?}
	The new finer model $\smplfine(\shape,\pose,\offsets, \shadingoffsets)$ consists of $N=110210$ vertices and $F=220416$ faces.
	%$\smplfine$ can be efficiently posed by changing parameters of the original SMPL model. 
	To recover the high-res smooth surface we calculate an initial set $\shadingoffsets_0 = \{\shadingoffset_0, \dots , \shadingoffset_e\}$ by minimizing
	\begin{equation}
	\small
	\operatorname*{arg\,min}_{\shadingoffsets} \Bigl(\mat{L}  \smplfine = \sum_{j \in \topology(i)} w_{ij} (\vertex_i - \vertex_j)\Bigr)
	\end{equation}
	where $\mat{L}$ is the Laplace matrix with cotangent weights $w_{ij}$ and $\topology(i)$ defines the neighbors around $\vertex_i$.
	
	\subsection{Medium-level body shape reconstruction}
	\label{sec:medium}
	In recent work, a pipeline to recover a subject's body shape, hair and clothing in the same setup as ours has been presented~\cite{alldieck2018video}.
	They first select a number of key-frames ($K \approx 120$) evenly distributed over the sequence and segment them into foreground and background using a CNN \cite{caelles2017oneshot}.
	Then they recover the 3D pose for each selected frame based on 2D landmarks \cite{cao2017realtime}. At the core of their method they transform the silhouette cone of every key-frame back into the canonical T-pose of the SMPL model using the inverse formulation of SMPL. This allows efficient optimization of the body shape independent of pose. We follow their pipeline and optimize for the subjects body shape in unposed space. However, we notice that the face estimation of \cite{alldieck2018video} is not accurate enough. This prevents us from further recovering fine-level facial features in the following steps, since precise face alignment is necessary for that. To this end, we propose a new objective for body shape estimation (dependency on parameters removed for clarity):
	\begin{equation}
	\operatorname*{arg\,min}_{\shape, \offsets} E_{\text{silh}} + E_{\text{face}} + E_{\text{regm}}
	\vspace{-1mm}
	\end{equation}
	The silhouette term $E_{\text{silh}}$ measures the distance between boundary vertices and silhouette rays. See \cite{alldieck2018video} for details and regularization $E_{\text{regm}}$.
	%
	%We follow their proposed method for consensus shape estimation as initialization to our method and add an important aspect.
	%Our medium-level body shape reconstruction is similar to \cite{alldieck2018video}.
	%
	%
	%To this end, we extend the shape optimization objective with a face alignment term $E_{\text{face}}$, which penalizes the distance between the 2D facial landmark detection and the 2D projection of 3D facial landmarks.
	%\TA{we need to make the point that the face is very important for a person's identity}
	%By adding face detection to the method we regularize the body model away from the SMPL mean shape towards the subject's individual facial features.
	The face alignment term $E_{\text{face}}$ penalizes the distance between the 2D facial landmark detections and the 2D projection of 3D facial landmarks.
	We use \added{OpenPose} \cite{simon2017hand} \deleted{trained for face key-point detection} to detect 2D facial landmarks for every key-frame.
	In order to incorporate the detections into the method, we establish a static mapping between landmarks and points on the mesh. Every landmark $\landmark$ is mapped to the surface via barycentric interpolation of neighboring vertices.
	%We define landmarks on the mesh surface by:
	%
	%\begin{equation}
	%\landmark(l, \vec{b}) = \sum_{i \in \face_l} b_i \vertex_i
	%\end{equation}
	%
	%where $\face$ is the set of vertices connected in a mesh face and $\vec{b}$ are barycentric coordinates.
	During optimization, we measure the point to line distance between the landmark $\landmark$ on the model and the corresponding camera ray $\ray$ describing the 2D landmark detection in unposed space:
	\begin{equation}
	\vspace{-1mm}
	\delta(\landmark, \ray) = \landmark \times \raycomp_n - \raycomp_m
	\end{equation}
	where $\ray = (\raycomp_m, \raycomp_n)$ is given in Plucker coordinates. 
	The face alignment term finally is: %that we add to the shape optimization objective in \cite{alldieck2018video} can be expressed as:
	\begin{equation}
	E_{\text{face}} = \sum_{l,r \in \landmarkmatches} w_l \rho(\delta(\landmark_l, \ray_r))
	\vspace{-1mm}
	\end{equation}
	where $\landmarkmatches$ defines the mapping between mesh points and landmarks, $w$ is the confidence of the landmark given by the CNN and $\rho$ is the Geman-McClure robust cost function. To speed up computation time, we use the coarse SMPL model formulation (Eq.~\ref{eq_smpl_coarse}) for the medium-level shape estimation.
	
	\subsection{Modeling fine-level surface details}
	\label{sec:fine}
	
	In Sec.~\ref{sec:medium}, we capture the medium-level details by globally integrating the silhouette information from all key-frames.
	Now our goal is to obtain fine-level surface details, which cannot be estimated from silhouette, based on shape-from-shading.
	Note that estimating shape-from-shading globally over all frames would lead to a smooth shape without details\added{, due to fabric movement and misalignments}. Thus, we first capture the details for a number of key-frames individually, and then incrementally merge the details into the model as new triangles become visible in a consecutive key-frame.
	We found that the number of key-frames can be lower than in the first step and choose $K=60$.
	Now we describe how to capture the fine-level details for a single key-frame $k$ based on shape-from-shading.
	To make this process robust, we estimate shading normals individually in a window around the key-frame and then jointly optimize for the surface.
	%Then we merge the normal estimates within the window.
	
	\textbf{Shape-from-shading:}
	For each frame, we first decompose the image into reflectance $\image_r$ and shading $\image_s$ using the CNN based intrinsic decomposition method of \cite{nestmeyer2017reflectanceFiltering}.
	The function $H_{\sh}$ calculates the shading of a vertex with spherical harmonic components $\sh$. 
	We estimate spherical harmonic components $\sh$ that minimize the difference between the simulated shading and the observed image shading $\image_s$
	%Using $H_{\sh}$, we can estimate the scene illumination %using spherical harmonics
	jointly for the given window of frames~\cite{sfs:Wu:CVPR2011}:
	\begin{equation}
	\operatorname*{arg\,min}_{\sh} \sum_{i \in \visvertices} \left| H_{\sh}(\normal_i) - \image_s(\projection \vertex_i) \right|,
	\vspace{-2mm}
	\end{equation}
	where  $\visvertices$ denotes the subset of visible vertices, i.e. the angle between the normal and the viewing direction is $0 < \alpha \leq \alpha_\text{max}$.
	$\projection$ is the projection matrix.
	Having the scene illumination and the shading for every pixel, we can now estimate auxiliary normals $\tilde{\normals} = \{\tilde{\normal}_0, \dots, \tilde{\normal}_N\}$ for every vertex per frame:
	\begin{equation}
	\operatorname*{arg\,min}_{\tilde{\normals}} E_\text{grad} + w_\text{lapn} E_\text{lapn}.
	\vspace{-2mm}
	\end{equation}
	The Laplacian smoothness term $E_\text{lapn} = \mat{L}\tilde{\normals}$ enforces the normals to be locally smooth.
	$E_\text{grad}$ penalizes shading errors by calculating the difference between the gradient between a shaded vertex and its neighbors $\topology$ and the image gradient at the projected vertex positions:
	\begin{equation}
	\small
	E_\text{grad} =  \sum_{i \in \visvertices} \sum_{j \in \topology(i) \cap \visvertices}  \left|\left| \Delta_{H_{\sh}}(\tilde{\normal}_i, \tilde{\normal}_j) - \Delta_{\image_s}(\projection \vertex_i, \projection \vertex_j) \right|\right|^2
	\end{equation}
	with $\Delta_f(a, b) = f(a) - f(b)$.
	
	\begin{figure}
		\centering
		\includegraphics[width=0.86\linewidth]{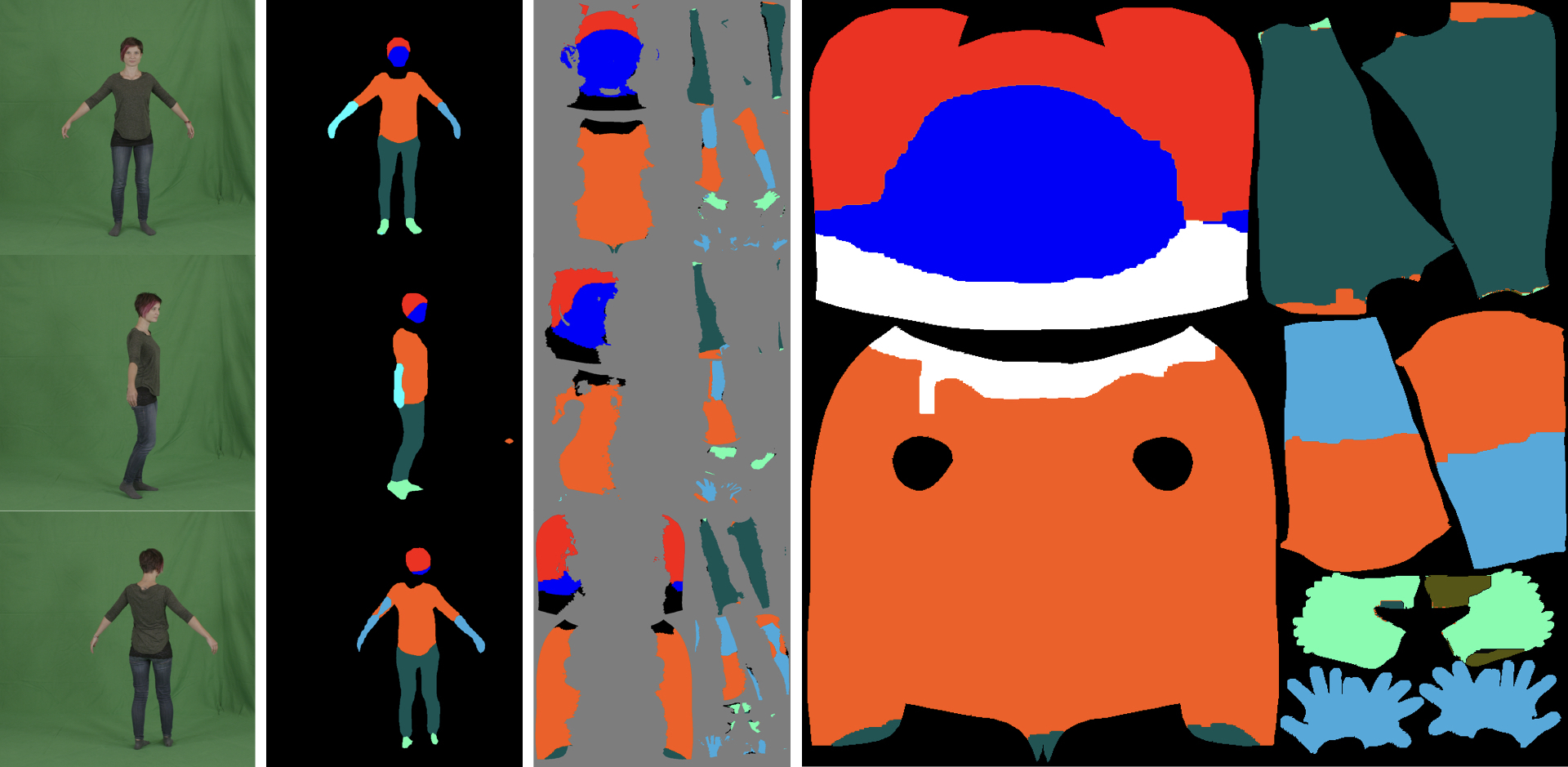}
		\caption{We calculate a semantic segmentation for every key frame. The semantic labels are mapped into texture space and combined into a semantic texture prior.}
		\label{fig:semantictexture}
		\vspace{-1mm}
	\end{figure}
	
	\vspace{2mm}
	\textbf{Surface reconstruction:}
	In order to merge information about all estimated normals within the window, we transform the normals back into the canonical T-pose using the inverse pose function of SMPL $\smpl^{-1}$. Then we optimize for the surface which explains the merged normals.
	Further, we include the silhouette term and face term of Sec.~\ref{sec:medium} to enforce the surface to be well aligned to the images.
	%Generally speaking we shift the focus of the consensus shape estimation to the current key-frame $k$ and at the same time estimate the interior surface by recovering the estimated normals.
	%Our objective function for surface reconstruction can be expressed as:
	%
	Specifically, we minimize:
	\begin{equation}
	\footnotesize
	\operatorname*{arg\,min}_{\offsets, \shadingoffsets} \sum_{j \in \set{C}}^{} \left( \lambda_j E_{\text{silh},j} + \lambda_j w_\text{face} E_{\text{face},j} \right) + w_{\text{sfs}}E_{\text{sfs}}  + E_{\text{regf}}
	\vspace{-1mm}
	\end{equation}
	with weights $w_\ast$  and $\lambda_j =1$ for $j=k$ and $\lambda_j<1$ otherwise.
	%The silhouette term $E_{\text{silh}}$ minimizes the distance between matches of boundary vertices and silhouette rays as described in~\cite{alldieck2018video}.
	$E_{\text{silh}}$ and $E_{\text{face}}$ are evaluated over a number of control frames $\set{C}$ and matches in $E_{\text{silh}}$ are limited to vertices in the original SMPL model.
	The shape-from-shading term is defined as:
	\begin{equation}
	\small
	E_\text{sfs} = \sum_{f = k - m}^{k + m} \sum_{i \in \visvertices} ||\normal_{i} - \tilde{\normal}^f_{i}||^2
	\end{equation}
	where $k$ is the current key-frame and $m$ specifies the window size, usually $m=1$. $\tilde{\normal}^f_{i}$ denotes the auxiliary normal of vertex $i$ calculated from frame $f$. All normals are in T-pose space. %(see \cite{alldieck2018video} for details about unposing)
	$E_{\text{regf}}$ regularizes the optimization as described in the following:
	\begin{equation}
	\footnotesize
	E_{\text{regf}} = w_{\text{match}}E_{\text{match}} + w_{\text{lap}}E_{\text{lap}} + w_{\text{struc}}E_{\text{struc}} + w_{\text{cons}}E_{\text{cons}}
	\label{eq_reg}
	\end{equation}
	$E_{\text{match}}$ penalizes the discrepancy between two neighboring key-frames.
	Specifically, for a perfect estimation, the following assumption should hold: When warping a key-frame into a neighboring key-frame based on the warp-field described by the projected vertex displacement, the warped frame and the target frame should be similar. 
	$E_{\text{match}}$ describes this metric: First we calculate the described warp. Then we calculate warping errors based on optical flow~\cite{brox2004high}. Based on the sum of the initial warp-field and the calculated error, we establish a grid of correspondences between neighboring key-frames. Every correspondence  $c$ should be explained by a particular point of the mesh surface. We first find a candidate for every correspondence:
	\begin{equation}
	\small
	\operatorname*{arg\,min}_{i \in \visvertices} \frac{\cos (\alpha^i_k) \delta(\vertex_i^k, \ray_c^k) + \cos (\alpha^i_j) \delta(\vertex_i^j, \ray_c^j)}{\cos (\alpha^i_k) + \cos (\alpha^i_j)}
	\label{cor_match}
	\end{equation}
	where $\alpha^i_k$ is the viewing angle under which the vertex $i$ has been seen in key-frame $k$ and $\ray_c^k$ is the projection ray of correspondence $c$ in posed space of key-frame $k$.
	Then we minimize point to line distance in unposed space:
	\begin{equation}
	E_{\text{match}} = \sum_{i,c \in \set{M}} \rho(\delta(\vertex_i, \ray_c))
	\vspace{-2mm}
	\end{equation}
	where $\set{M}$ is the set of matches established in Eq.~\ref{cor_match}.
	
	The remaining regularization terms of Eq.\ref{eq_reg} are as follows:
	$E_{\text{lap}}$ is the Laplacian smoothness term with anisotropic weights~\cite{sfs:Wu:CVPR2011}.
	$E_{\text{struc}}$ aims to keep the structure of the mesh by pruning edge length variations.
	$E_{\text{cons}}$ prunes large deviations from the consensus shape.
	
	We optimize using a \emph{dog-leg} trust region method using the chumpy autodifferentiation framework. We alternate minimizing and finding silhouette point to line correspondences. Regularization is reduced step-wise.
	
	\begin{figure}
		\centering
		\includegraphics[width=\linewidth]{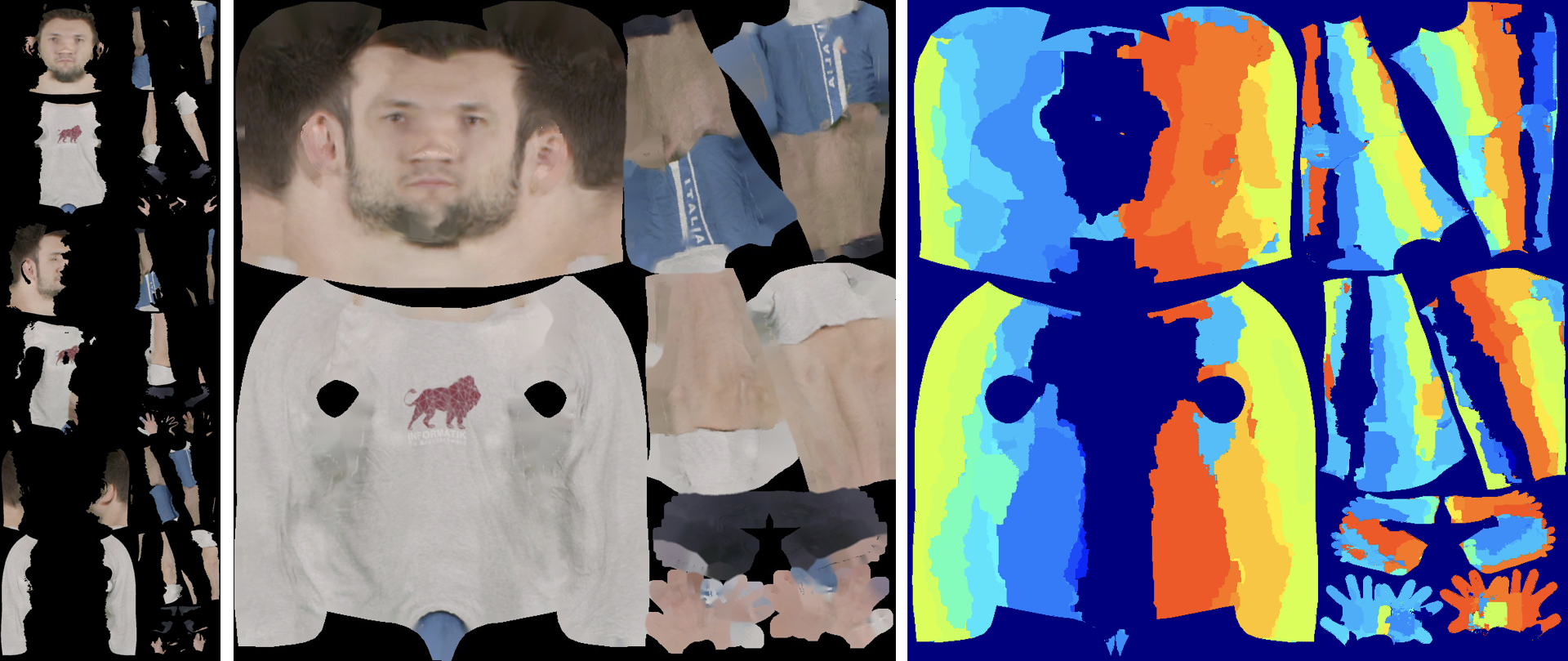}
		\caption{Based on part textures from key frames (left), we stitch a complete texture using  graph-cut based optimization. The associated key frames for each texel are shown as colors on the right.}
		\label{fig:texturestitch}
		\vspace{-1mm}
	\end{figure}

	\subsection{Texture generation}
	A high quality texture image is an essential component for a realistic virtual character, since it can describe the material properties that cannot be modeled by the surface geometry.
	In order to obtain a sharp and seamless texture, we solve the texture stitching on a per texel level (Fig.~\ref{fig:texturestitch}), in contrast to that on a per face level as in other works~\cite{lempitsky2007seamless}.
	In other words, our goal is to color each pixel in the texture image with a pixel value taken from one out of $K$ key-frames. %, rather than estimate the per face mapping between the texture image and input images.
	However, this makes the scale of our problem much larger, and therefore does not allow us to perform global optimization.
	To this end, we propose a novel texture merging method based on graph cut, which translates our problem to a series of binary labeling subproblems that can be efficiently solved.
	Furthermore, meshes and key-frames are not perfectly aligned. %This means that mesh body parts might cover other parts or background in the image.
	To reduce color spilling and artifacts caused by misalignments, we compute a semantic prior before stitching the final texture (Fig.~\ref{fig:semantictexture}).
	
	%Specifically, we first generate a partial texture $\mat{U}_k$ for every key-frame $k$, and then calculate the complete texture by incrementally merge the partial textures with graph cut.
	
	\textbf{Partial texture generation:}
	For every key-frame, we first project all visible surface points to the frame and write the color at the projected position into the corresponding texture coordinates.
	In order to factor out the illumination in the texture images, we \emph{unshade} the input images by dividing them with the shading images as used in Sec.~\ref{sec:fine}.
	The partial texture calculation can easily be achieved using the OpenGL rasterization pipeline.
	Apart from the partial color texture image, we calculate two additional texture maps for the merging step, i.e. the viewing-angle map and the semantic map.
	For the viewing-angle map, we compute the viewing angle $\alpha^t_k$ under which the surface point $t$ has been seen in key-frame $k$.
	
	\textbf{The semantic prior} is generated by re-projecting the human semantic segmentation to the texture space.
	Specifically, we first calculate a semantic label for every pixel in the input frames using a CNN based human parsing method~\cite{liang2015humanparsing}.
	Each frame is segmented into 10 semantic classes such as \emph{hair}, \emph{face}, \emph{left leg} and \emph{upper clothes}.
	Then the semantic information of all frames is fused into the global semantic map by minimizing for labeling $\vec{x}$:
	\begin{equation}
	\vspace{-2mm}
	\small
	\operatorname*{arg\,min}_{\vec{x}} \sum_{t = 0}^{T} \varphi_t(x_t) + \sum_{t,q \in \neighbors} \psi(x_t, x_q)
	\label{graphcutsegment}
	\end{equation}
	\begin{equation}
	\small
	\varphi_t(x_t) = 1 - \frac{\sum_{k=0}^K X_k(\cos^2 \alpha^t_k)}{K}
	\end{equation}
	Here $\varphi$ is the energy term describing the compatibility of a label $x$ with the texel $t$, where $X_k$ returns the given value if the texel was labeled with $x$ in view $k$ and $0$ otherwise.
	$\psi$ gives the label compatibility of neighboring texels $t$ and $q$.
	We solve Eq.~\ref{graphcutsegment} by multi-label graph-cut optimization with alpha-beta swaps~\cite{boykov2001fast}.
	While constructing the graph, we connect every texel not only with its neighbors in texture space but with all neighbors on the surface.
	In particular this means texels are connected across texture seams.
	%A semantic label for every texel alone cannot be used a prior for texture completion.
	To have a strong prior for the texture completion, we calculate Gaussian mixture models (GMM) of the colors in HSV space per label using the part-textures and corresponding labels.
	
	\textbf{Texture merging:}
	Next, we calculate the complete texture by merging the partial textures. While keeping the same graph structure, the objective function is:
	\vspace{-2mm}
	\begin{equation}
	\small
	\operatorname*{arg\,min}_{\vec{u}} \sum_{t = 0}^{T} \theta_t(u_t) + \sum_{t,q \in \neighbors} \eta_{t, q}(u_t, u_q)
	\label{texture_stitch}
	\vspace{-2mm}
	\end{equation}
	where the labeling $\vec{u}$ assigns every texel to a partial texture $k$.
	The first term seeks to find the best image for each texel:
	\begin{align}
	\small
	\theta_t(k) = w_\text{vis} \sin^2 \alpha_k^t + w_\text{gmm} m(\mat{U}_k^t, x_t) \nonumber \\
	+ w_\text{face} d(\mat{U}_k^t)  + w_\text{silh} E_{\text{silh}, k}
	\end{align}
	with weights $w_\ast$. $m$ returns the Mahalanobis distance between the color value for $t$ in part-texture $k$ given the semantic label $x_t$. $d$ calculates the structural dissimilarity between the first and the given key-frame. $d$ is only evaluated on texels belonging to the facial region and ensures consistent facial expression over the texture.
	
	The smoothness-term $\eta$ ensures similar colors for neighboring texels.
	For neighboring texels assigned to different key-frames $u_t \neq u_q$, while belonging to the same semantic region $x_t = x_q$, $\eta_{t, q}$ equals the gradient magnitude between the texel colors $||\mat{U}_{u_t}^t - \mat{U}_{u_q}^q||$.
	
	Since the number of combinations in $\eta$ is very high, it is computationally not feasible to solve Eq.~\ref{texture_stitch} as a multi label graph-cut problem. 
	Thus, we propose the following strategy for an approximate solution:
	We convert the multi-label problem to a binary labeling decision $b \in \{\text{\emph{update}}, \text{\emph{keep}}\}$.
	We initialize the texture with $\mat{M} = \mat{U}_0$.
	Then we randomly choose a key-frame $k$ and test it against the current solution.
	The likelihood of selecting a key-frame is inversly proportional to its remaining silhouette error $E_{\text{silh}, k}$ in order to favor well-aligned key-frames.
	Further, $\eta$ is approximated with:
	\begin{equation}
	\footnotesize
	\eta_{t,q} = 
	\begin{cases}
	\max(||\mat{M}^t - \mat{U}^q_k||, ||\mat{M}^q - \mat{U}^t_k||),& \text{if } b_t \neq b_q \land x_t = x_q\\
	0,   & \text{otherwise}
	\end{cases}
	\end{equation}
	Convergence is usually reached between $2K$ to $3K$ iterations. Finally, we cross-blend between different labels to reduce visible seams. \added{The run-time per iteration on $1000\times1000$ px with Python code using a standard  graph cut library is ${\sim}2$~sec. No attempts for run-time optimization have been made.}

	\section{Experiments}
	\label{sec:experiments}
	
	We evaluate our method on two publicly available datasets: The People-Snapshot dataset~\cite{alldieck2018video} and the dataset used in~\cite{Bogo:ICCV:2015}. 
	% and one synthetic dataset.
	To validate the perceived quality of our results we performed a user study.
	
	\begin{figure}\centering\offinterlineskip%
		\includegraphics[width=0.33\columnwidth]{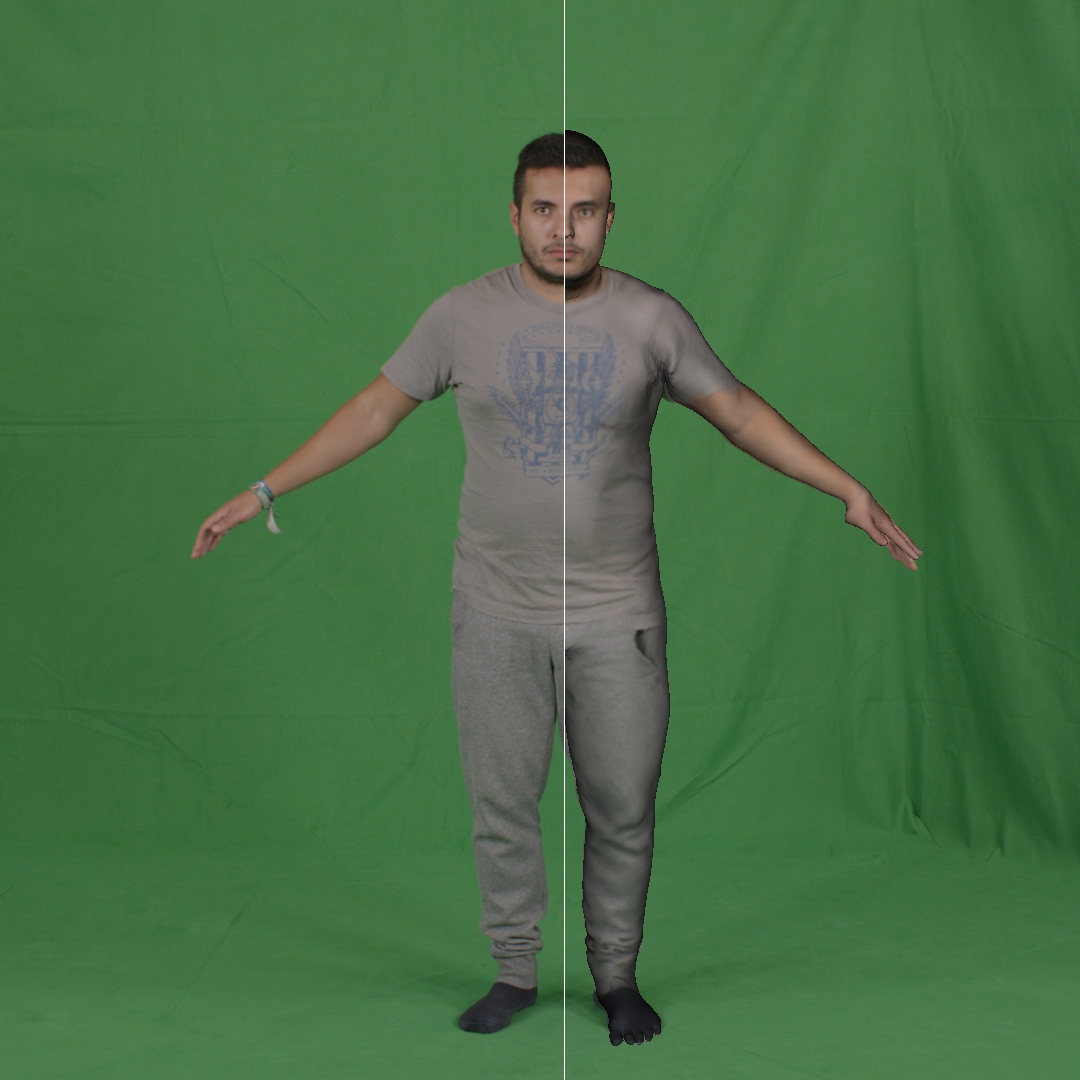}%
		\includegraphics[width=0.33\columnwidth]{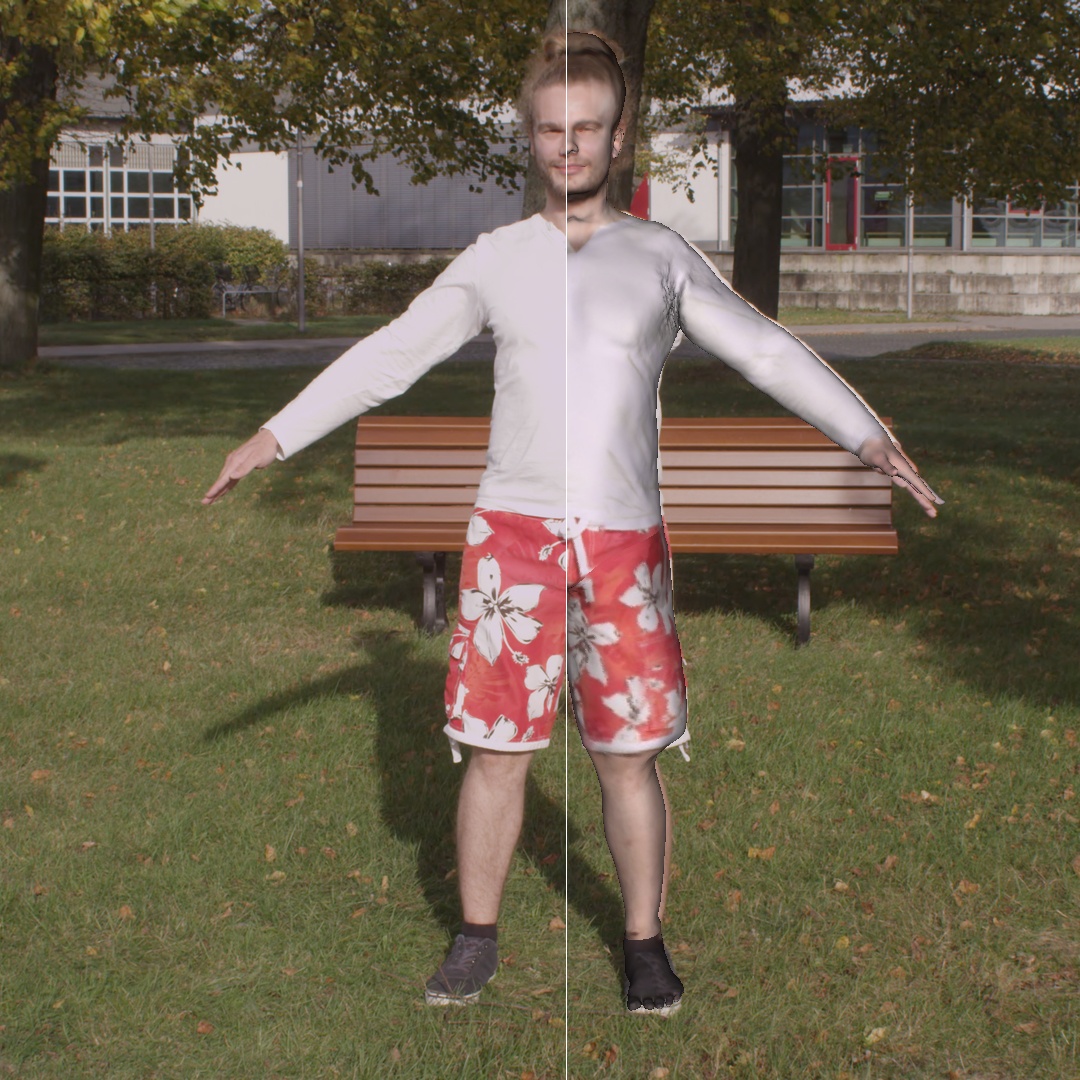}%
		\includegraphics[width=0.33\columnwidth]{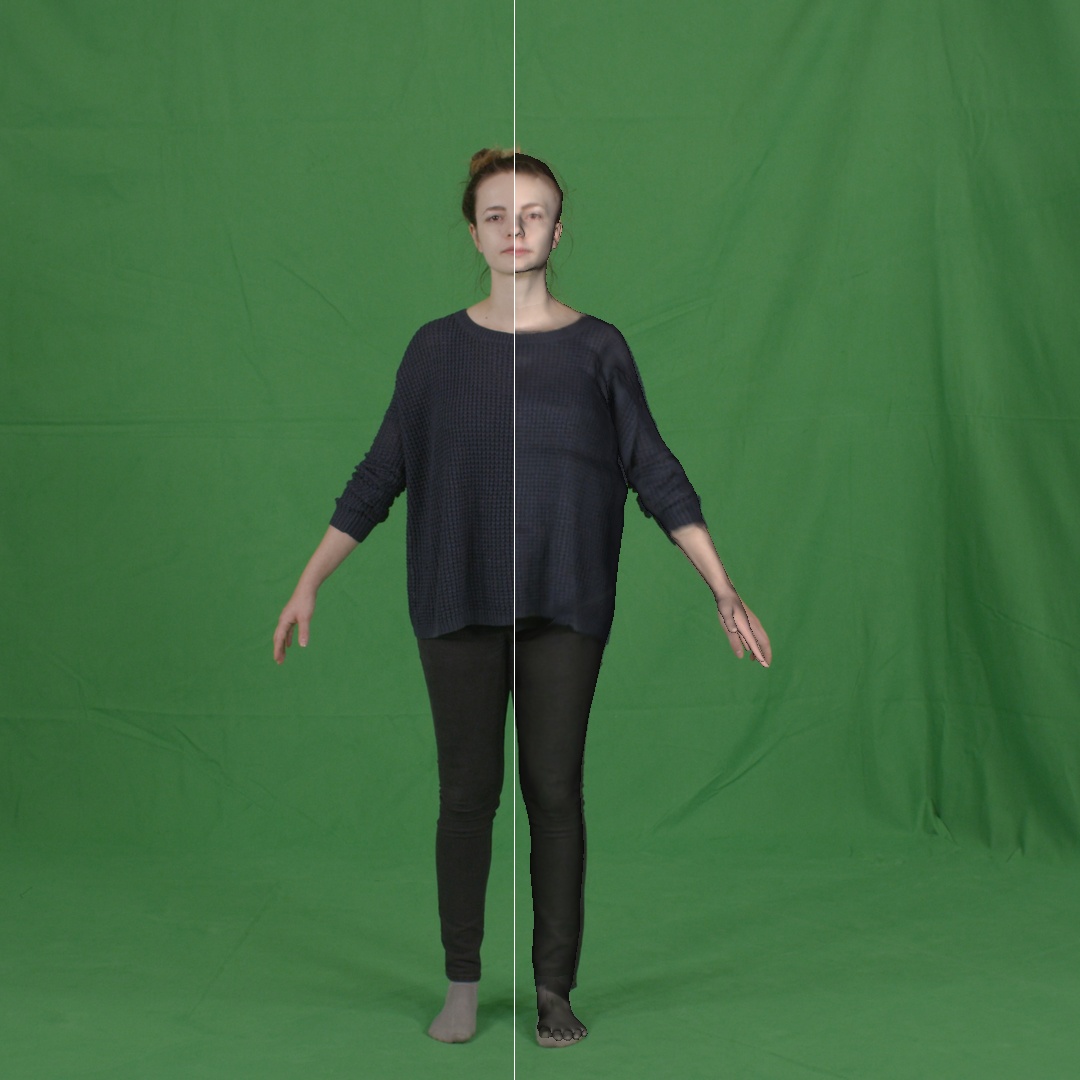}\\%
		\includegraphics[width=0.33\columnwidth]{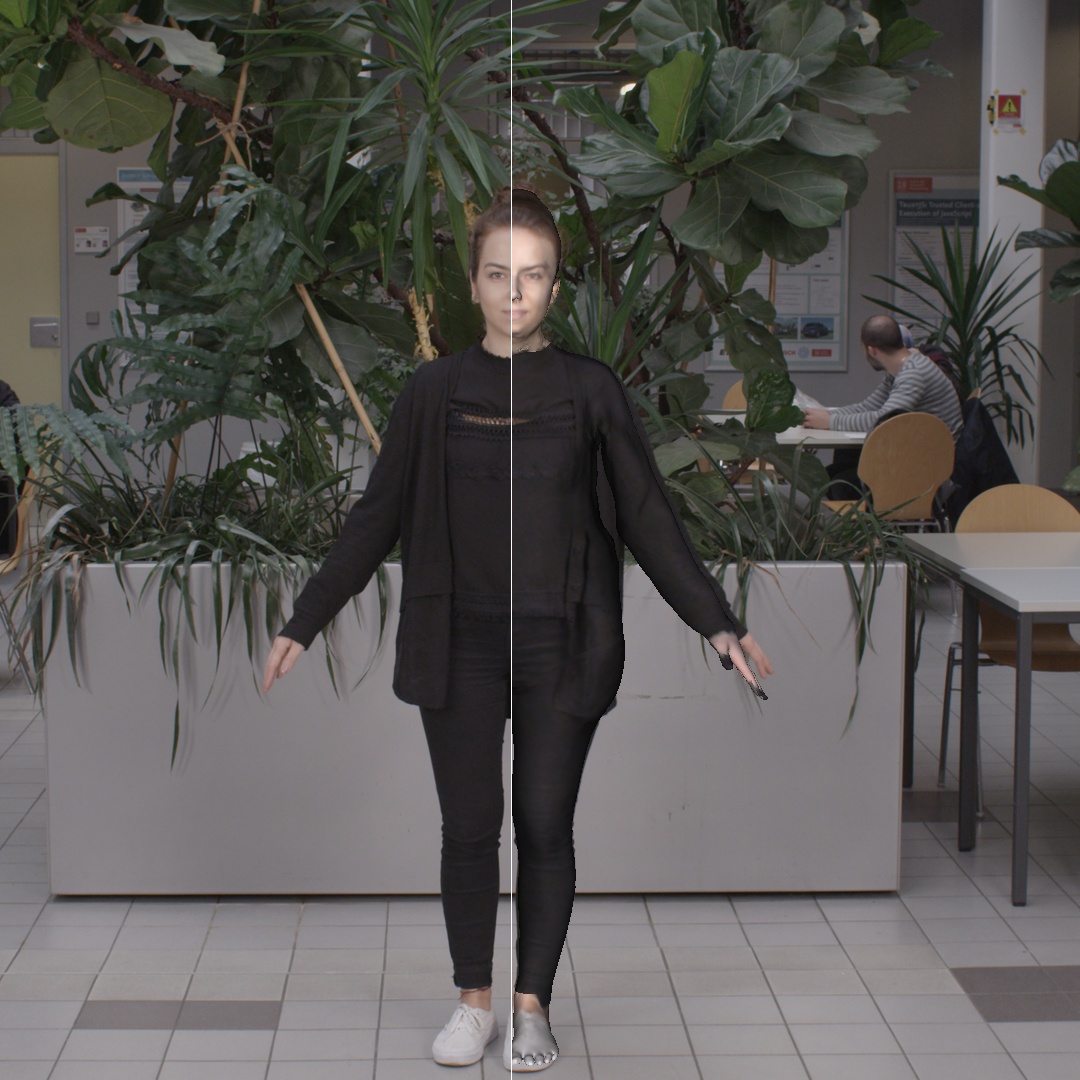}%
		\includegraphics[width=0.33\columnwidth]{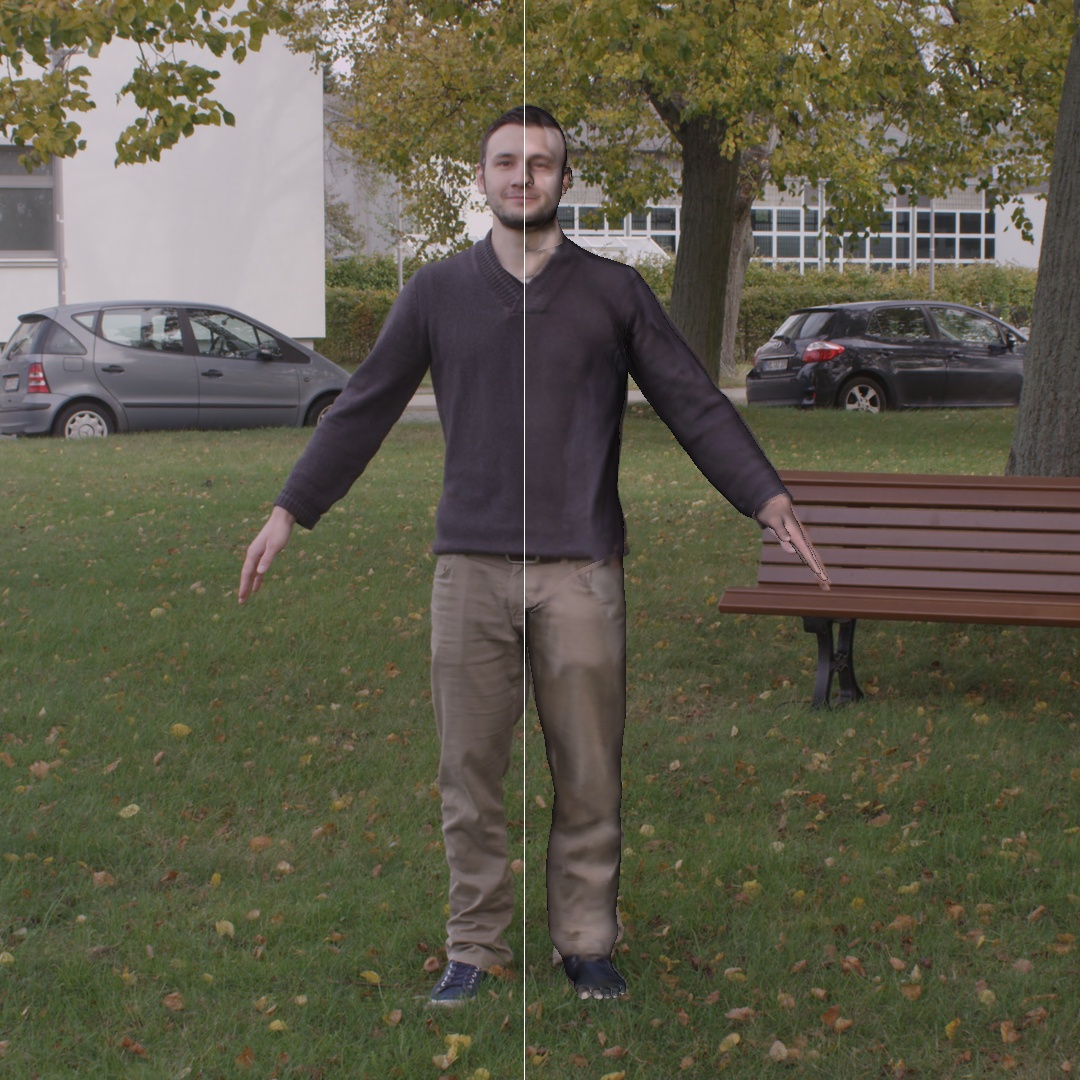}%
		\includegraphics[width=0.33\columnwidth]{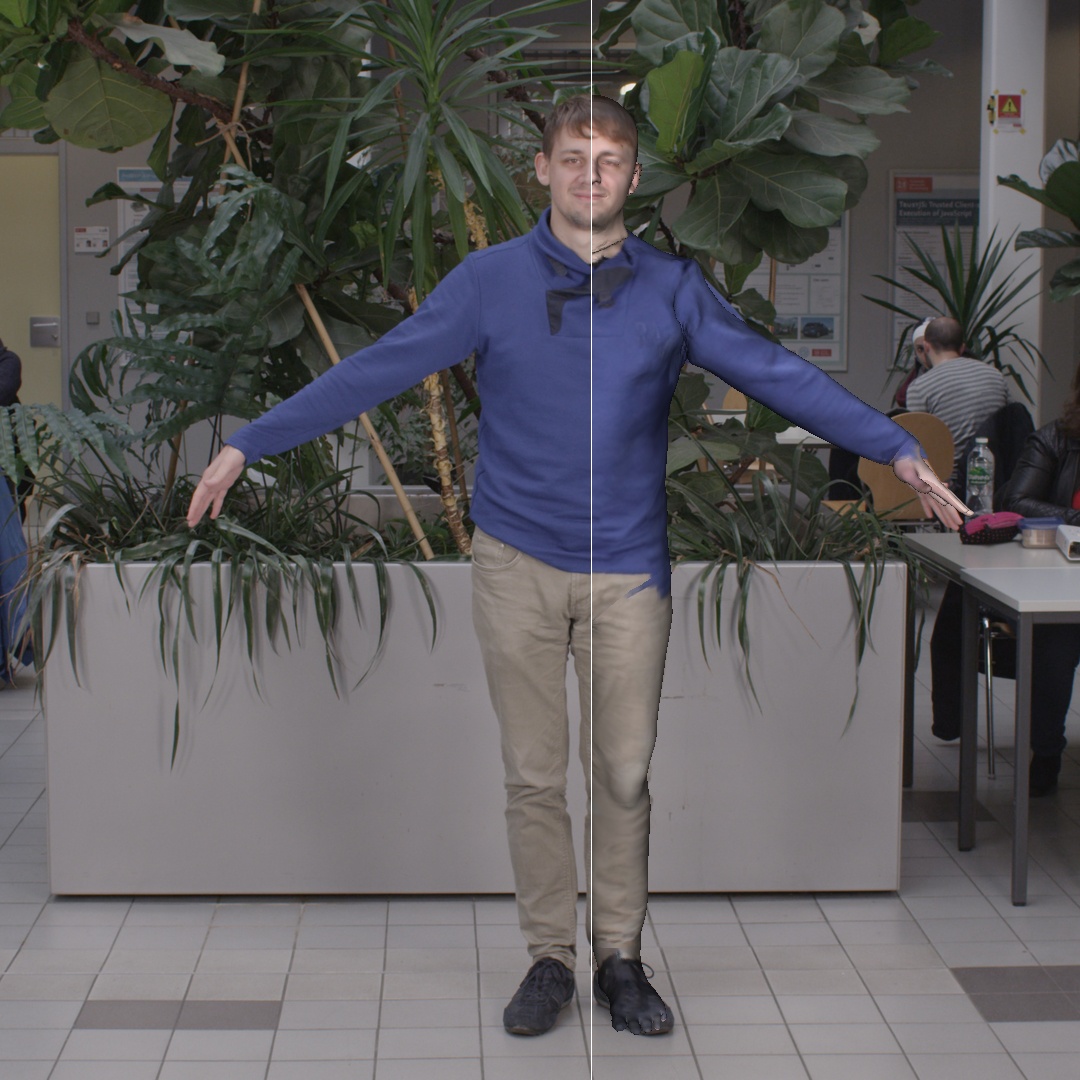}
		\begin{minipage}{\columnwidth}
			\footnotesize
			\centering
			\color{white}
			\vspace{-58mm}
			a)\hspace{10mm} ~b)\hspace{0.66\columnwidth}
		\end{minipage}
		\vspace{2mm}
		\caption{Side-by-side comparisons of our reconstructions (b) and the input frame (a). As can be seen from (b), our method closely resembles the subject in the video (a).}
		\label{fig:sidebyside}
		\vspace{-3mm}
	\end{figure}

	\subsection{Qualitative results and comparisons}
	\label{sec:qualiresults}
	We compare our method to the recent method of~\cite{alldieck2018video} on their People-Snapshot dataset. The approach of~\cite{alldieck2018video} is the only other monocular 3D person reconstruction method.
	%We provide qualitative comparison of our method against the state-of-the-art on their People-Snapshot dataset \cite{alldieck2018video}.
	The People-Snapshot dataset consists of 24 sequences of different subjects rotating in front of the camera while roughly holding an A-pose.
	In Fig.~\ref{fig:sidebyside}, we show some examples of our reconstruction results, which precisely overlay the subjects in the image.
	%The level of detail of our reconstructions ensures very high similarity between input frames and rendered avatars.
	Note that the level of detail of the input images is captured by our reconstructed avatars.
	In Fig.~\ref{fig:cvprcmp}, we show side-by-side comparison to~\cite{alldieck2018video}. Our results (right) reconstruct the face better and preserve many more details, e.g.\ clothing wrinkles and t-shirt stamps.
	
\begin{figure}
		\centering
		\includegraphics[width=0.7\linewidth]{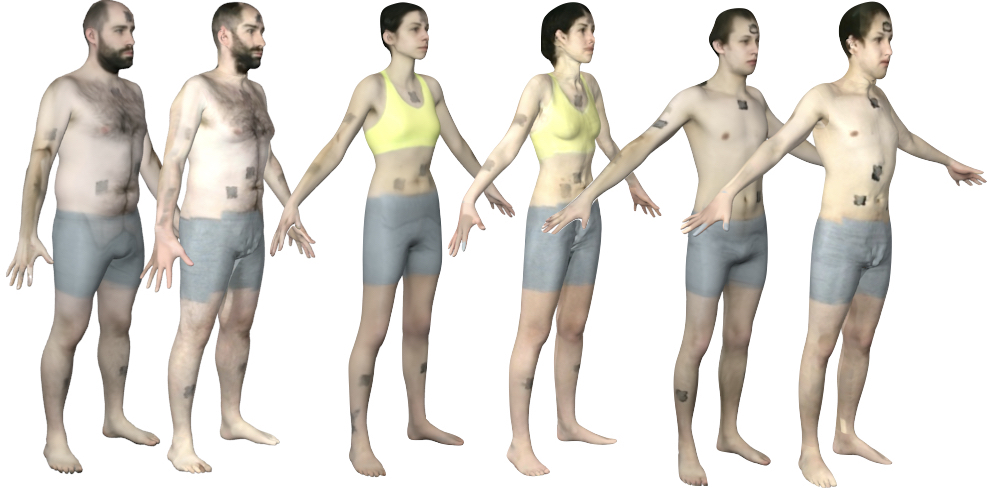}
		\begin{minipage}{1.0\textwidth}
			\footnotesize
			\vspace{-7mm}
			\hspace{18mm}a)\hspace{6mm}b)
		\end{minipage}
		\vspace{-8mm}
		\caption{Our results (b) in comparison against the RGB-D method \cite{Bogo:ICCV:2015} (a). Note that the texture prior has not been used (see Sec.~\ref{sec:qualiresults}).}
		\label{fig:kinectcap}
		%\vspace{-2mm}
	\end{figure}

	Additionally, we compare against the state-of-the-art RGB-D method~\cite{Bogo:ICCV:2015}, also using their dataset of people in minimal clothing\footnote{The deep learning based segmentation~\cite{gong2017look} only works for fully clothed people so we had to deactivate the semantic prior in this dataset.}.
	While their method relies on depth data, we only use the RGB video which makes the problem much harder. 
	Despite this, as shown in Fig.~\ref{fig:kinectcap}, our results are comparable in quality to theirs.
	
	%Note that the semantic segmentation needed for the texture prior could not be computed for subjects wearing minimal clothing.
	
	%Finally we show results on a synthetic dataset. The DynamicFAUST dataset~\cite{dfaust:CVPR:2017} contains scans of subjects in motion. The subjects vary a lot in height in weight. We render meshes of all 9 subjects performing the \emph{Hip} movement, such as the camera rotates in full circle over the lengths of the sequence. Since DynamicFAUST contains no textures, we simply shade the meshes. Renderings of shaded meshes and real images are structurally different, hence the face detector is not working for the given setup. We store the ground truth projected visible face landmarks, while rendering the sequence, and use them for our reconstructions. \TA{add image and discussion}
	
	\begin{figure}
		\centering
		\includegraphics[width=1\linewidth]{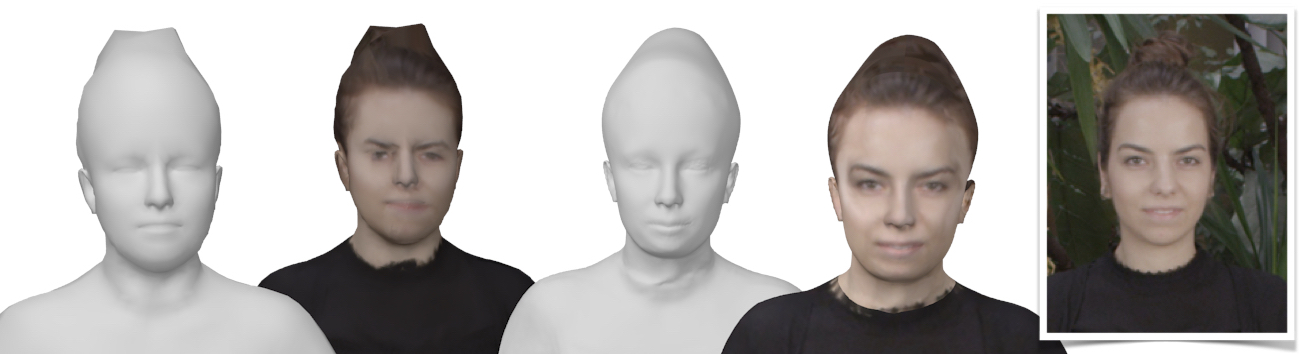}
		\caption{In comparison to the method of \cite{alldieck2018video} (left), the faces in our results (right) have finer details in the mesh and closely resemble the subject in the photograph.}
		\label{fig:face}
		\vspace{-3mm}
	\end{figure}

	\subsection{\added{Face similarity measure}}
\begin{figure}\centering\offinterlineskip
		\includegraphics[width=1\linewidth]{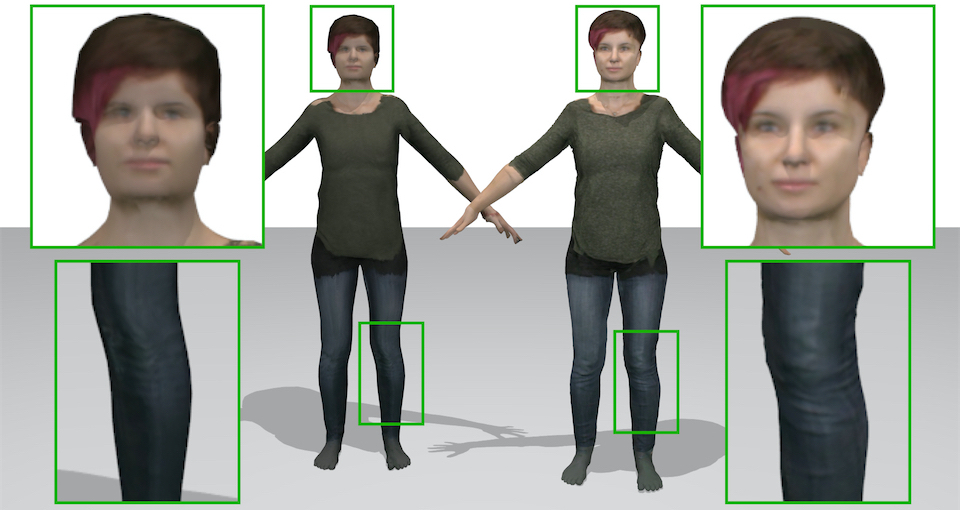}\\%
		\includegraphics[width=1\linewidth]{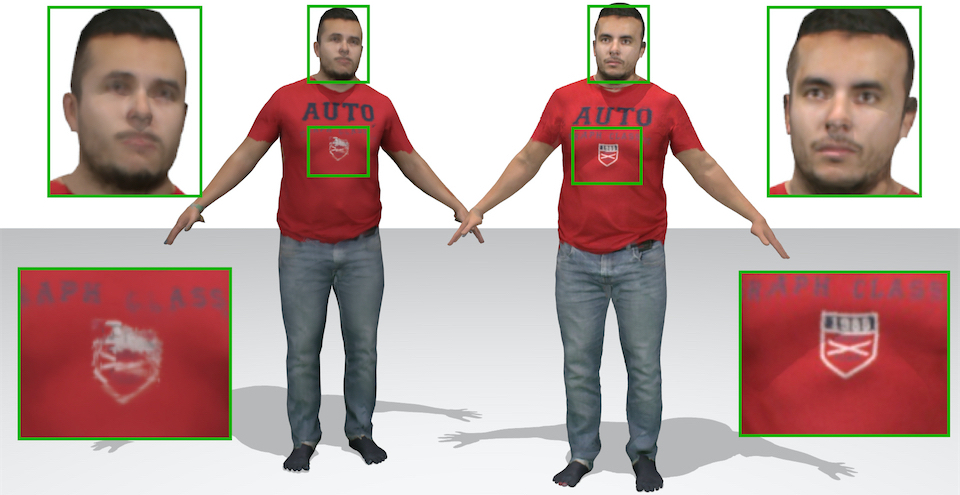}\\%
		\includegraphics[width=1\linewidth]{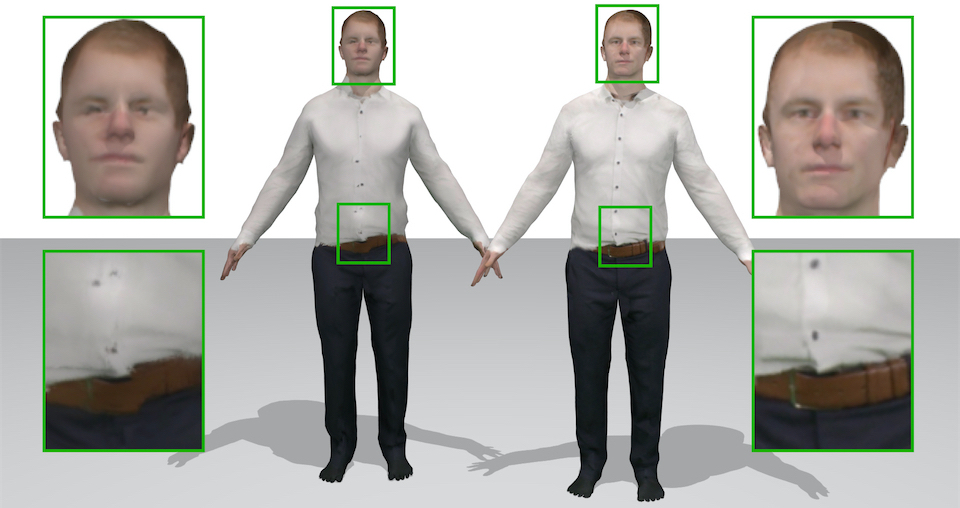}%
		\vspace{1.5mm}
		\caption{In comparison to the method of \cite{alldieck2018video} (left), our results (right) look much more natural and have finer details.}
		\label{fig:cvprcmp}
		%\vspace{-2mm}
	\end{figure}

	One goal of our method was to preserve the individual appearance of subjects in their avatars. Since the face is crucial for this, we leverage facial landmarks detections and shape-from-shading. As seen in Fig.~\ref{fig:cvprcmp} our method adds a significant level of detail to the facial region in comparison to state-of-the-art. In Fig.~\ref{fig:face} we show the same comparison also for untextured meshes. Our result closely resembles the subject in the photograph.
	\added{To further demonstrate the effectiveness of our method for face similarity preservation, we perform the following experiment: FaceNet~\cite{schroff2015facenet} is a deep network, that is trained to map from face images to an Euclidean space where distance corresponds to face similarity. We use FaceNet trained on the CASIA WebFace dataset~\cite{yi2014learning} to measure the similarity between photos of the subjects in the People-snapshot dataset and their reconstructions. Two distinct subjects in the dataset have a mean similarity distance of $1.33\pm0.13$. Same subjects in different settings differ by $0.55\pm0.18$. Our reconstructions feature a mean distance of $0.99\pm0.11$ to their photo counterparts. Reconstructions of~\cite{alldieck2018video} perform significantly worse with a mean distance of $1.09\pm0.15$. While our reconstructions can be reliable identified using FaceNet, reconstructions of~\cite{alldieck2018video} have a similarity distance close to a distance of distinct people, making them less likely to be identified correctly.}
	
	\subsection{Ablation analysis}
	In the following we qualitatively demonstrate the effectiveness of further design choices of our method.
	
\begin{figure}
		\centering
		\includegraphics[width=0.78\linewidth]{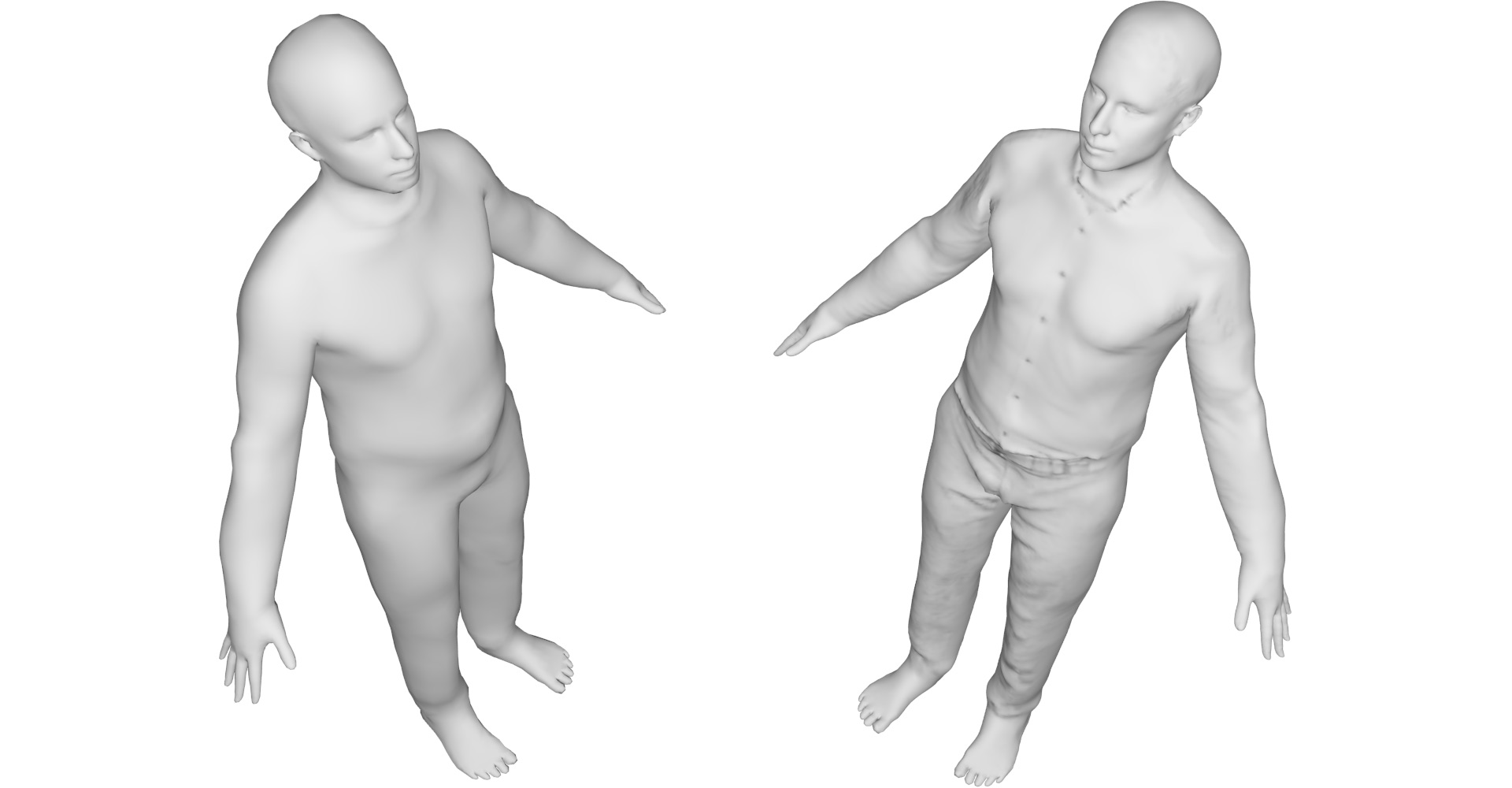}
		\vspace{-0.5mm}
		\caption{Comparison of a result of our method before (left) and after (right) applying shape-from-shading based detail enhancing.}
		\label{fig:sfs}
		%\vspace{-2mm}
	\end{figure}

	\textbf{Shape-from-shading:} In order to render the avatars under different illuminations, detailed geometry should be present in the mesh. In Fig.~\ref{fig:sfs}, we demonstrate the level of detail added to the meshes by shape-from-shading. While the mesh on the left only describes the low-frequency shape, our refined result on the right contains fine-grained details such as wrinkles and buttons. %\GPM{Promise more on the video?}
	
\begin{figure}
		\centering
		\includegraphics[width=0.72\linewidth]{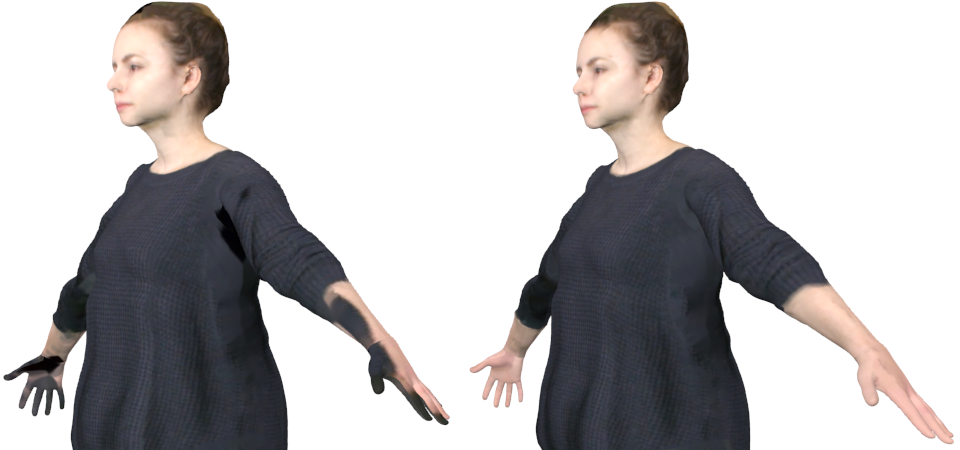}
		\vspace{-0.5mm}
		\caption{The semantic prior for texture stitching successfully removes color spilling (left) in our final texture (right).}
		\label{fig:texprior}
		\vspace{-3mm}
	\end{figure}

	\textbf{Influence of the texture prior:} In Fig.~\ref{fig:texprior} we show the effectiveness of the semantic prior for texture stitching. While the texture on the left computed without the prior contains noticeable color spills on the arms and hands, the final texture on the right contains no color spills and less stitching artifacts along semantic boundaries. %\GPM{texture prior or semantic map?}
	
	\subsection{User study}
	\label{user_study}
	Finally, we conducted a user study in order to validate the visual fidelity of our results. Each participant was asked four questions about 6 randomly chosen results out of the 24 reconstructed subjects in People-Snapshot dataset. 
	The avatars shown to each participant and the questions asked were randomized. In every question the participants had to decide between our method, and the method of \cite{alldieck2018video}. The four question were: 
	\vspace{-1mm}
	\begin{itemize} 
		\setlength\itemsep{-0.3em}
		\item[$-$] Which avatar preserves the identity of the person in the image better? (\emph{identity})
		\item[$-$] Which avatar has more detail? (\emph{detail})
		\item[$-$] Which avatar looks more real to you? (\emph{realism})
		\item[$-$] Which avatar do you like better? (\emph{preference})
	\end{itemize}
	%\vspace{-0.5mm}
	%
	\added{We presented the users renderings of the meshes in consistent pose and illumination. The users were allowed to zoom into the images.}
	At questions \emph{identity} and \emph{realism} we showed the participants either textured or untextured meshes.
	\added{For \emph{identity} comparison we additionally showed a photo of the subject next to the renderings.} When asking for \emph{detail} we only showed untextured meshes, and when asking for \emph{preference} we only showed textured results.
	Additionally, we asked for the level of experience with 3D data (\emph{None}, \emph{Beginner}, \emph{Proficient}, \emph{Expert}). %In total we interviewed 74 participants in an online survey, covering the whole range of expertise.
	74 people participated in our online survey, covering the whole range of expertise.
	
	The results of the study are summarized in Table~\ref{tab:study}. The participants clearly preferred our results in all scenarios over current state-of-the-art. 
	Admittedly, when asked about identity preservation in untextured meshes, users preferred our method, but this time only $65.70\%$. 
	%
	%The only outlier is the question about \emph{identity}, when showing untextured meshes, with only $65.70\%$ votes for our method. 
	%We propose the hypothesis that people are not used to looking at untextured meshes and are therefore not able to retrieve the necessary information to answer the question. 
	Further inspection of the results shows that users with high experience with 3D data think our method preserves the identity better with $90.48\%$ versus $60.49\%$ for novice users.
	%To support the hypothesis we look at the results grouped by experience of the participants. 
	%Here a fairly clear picture emerges: While participants with no experience in 3D graphics chose our method only $60.49\%$ of the times, experts clearly acknowledge that our method preserves the identity better with $90.48\%$ votes for our method. 
	We hypothesize that unexperienced users find it more difficult to recognize people from 3D meshes without textures.
	Most importantly, by a large margin, our results are perceived as more realistic (92.27\%), preserve more details (95.72\%) and where preferred 89.64\% of the times.  
	
	%\GPM{anyone minimally intelligent will see that if you have 60 vs 90 and you had 65, there are very few samples in the 90. Also I would not end the study with this negative result. }
	
	\begin{table}
		\centering
		\footnotesize
		%\begin{tabular}{c|c|c}
		%	& Textured Avatars & Untextured Avatars \\ 
		%	\hline \hline
		%	Identitiy & 83.12 \% & 65.70 \% \\ 
		%	Details & - & 95.72 \% \\ 
		%	Realism & 92.27 \% & 89.73 \% \\ 
		%	Preference & 89.64 \% & - \\ 
		%\end{tabular} 
		\begin{tabular}{c|c|c|c|c}
			& Identitiy & Details & Realism & Preference \\ 
			\hline \hline
			Textured Avatars & 83.12 \% & -  & 92.27 \% &  89.64 \% \\ 
			Untextured Avatars & 65.70 \% & 95.72 \% & 89.73 \% & -\\ 
		\end{tabular}%
		\vspace{1mm}
		\caption{Results of the user study. Percentage of answers where users preferred our method over \cite{alldieck2018video}. We asked for four different aspects. See Sec.~\ref{user_study} for details.}
		\label{tab:study}
		\vspace{-1mm}
	\end{table}

	\section{Discussion and Conclusion}
	\label{sec:conclusion}
	
	We have proposed a novel method to create highly detailed personalized avatars from monocular video.
	We improve over the state-of-the-art in several important aspects:
	Our optimization scheme allows to integrate face landmark detections and shape-from-shading from multiple frames.
	Experiments demonstrate that this results in better face reconstruction and better identity preservation. This is also confirmed by our user study, 
	which shows that people think our method preserves identity better 83.12\% of the times, and capture more details 95.72\% of the times.
	
	We introduced a new texture stitching binary optimization, which allows us to efficiently merge the appearance of multiple frames into a single coherent texture. 
	The optimization includes a semantic texture term that incorporates appearance models for each semantic segmentation part. 
	%Results demonstrate that this reduces color spilling from skin to clothing or viceversa, which is a common artifact. 
	Results demonstrate that the common artifact of color spilling from skin to clothing or viceversa gets reduced.
	%The result are avatars that closely resemble the subjects in the video.
	
	We have argued for a method to capture the subtle, but very important details to make avatars look \emph{realistic}. 
	Indeed \emph{details matter}, the user study shows that users think our results are more realistic than the state of the art 92.7\% of the times, and prefer our avatars 89.64\% of the times. 
	
	Future work should address capture of subjects wearing clothing with topology different from the body, including skirts and coats.
	Furthermore, to obtain full texturing, subjects have to be seen from all sides -- it may be possible to infer occluded appearance using sufficient training data. 
	Another avenue to explore is reconstruction in an un-cooperative setting, e.g.\ from online videos of people.  
	
	%Finally, semantic human parsing is yet not stable enough to serve as a texture prior in all possible scenarios.
	
	%Our method enables for the first time highly detailed and identity preserving human avatar creation from \emph{monocular video}. 
	Having cameras all around us, we can now serve the growing demand for personalized avatars in virtual and augmented reality applications e.g.\ in the fields of entertainment, communication or e-commerce.
	%-------------------------------------------------------------------------
	
	\vfill
	\noindent
	\footnotesize{\textbf{Acknowledgments:}
		The authors gratefully acknowledge funding by the German Science Foundation from project DFG MA2555/12-1.
	}
	\newpage
	{\small
		\bibliographystyle{ieee}
		\bibliography{egbib}
	}

\end{document}